\documentclass[preprint,12pt,review]{elsarticle}

\usepackage{amssymb,amsmath,paralist}
\usepackage{xfrac}
\usepackage{lipsum}
\usepackage[english]{babel}
\usepackage[utf8]{inputenc}
\usepackage{graphicx,xcolor,adjustbox}
\usepackage[font=small,labelfont=bf,tableposition=top]{caption}
\usepackage[font=footnotesize]{subcaption}
\usepackage{amsfonts}
\usepackage{mathtools}
\usepackage{etoolbox}
\usepackage{mdframed}
\usepackage{filecontents}
\usepackage{enumitem}

\usepackage{algorithm}
\usepackage{listings}
\usepackage{algpseudocode}

\usepackage{bbm}
\usepackage{multirow}
\usepackage{rotating}
\usepackage{float}
\usepackage{subfiles}

\usepackage{lineno}

\usepackage{nicematrix}
\usepackage[all,cmtip]{xy}
\usepackage{tikz}
\usetikzlibrary{calc,fit}
\usepackage{float}
\usepackage{scrextend}
\usepackage[]{upgreek}
\usepackage{tabularx}
\usepackage{booktabs}

\newcommand{\bfs}{\bfseries}

\newcommand{\re}{\mathbb{R}}
\newcommand{\nn}{\mathbb{N}}

\newcommand{\prob}[1]{\mathbb{P}\lpar #1 \rpar}

\newcommand{\lpar}{\left ( }
\newcommand{\rpar}{ \right )}
\newcommand{\lset}{\left\lbrace}
\newcommand{\rset}{\right\rbrace }
\newcommand{\lbr}{\left [  }
\newcommand{\rbr}{  \right ] }

\newcommand{\algName}{Stochastic Mean Shift\ }
\newcommand{\msop}{\mathcal{S}}  
\newcommand{\shv}{\mathbf{m}_{h}} 
\newcommand{\kde}{D} 
\newcommand{\shortname}{SMS\,\,}
\newcommand{\mshift}{MS\,\,}
\newcommand{\truedensity}{f_{X}}
\newcommand{\bms}{BMS\,\,}
\newcommand{\point}{\mathbf{x}}
\newcommand{\state}{\mathcal{X}}

\newcommand{\ridx}[1]{i_{#1}}
\newcommand{\neighb}[1]{N_{h} (#1)}

\newcommand{\SmallSubg}[1]{A_{#1}}

\newtheorem{theorem*}{{\bfs \text{Theorem}}}
\newtheorem{def*}{{\bfs \text{Definition}}}
\newtheorem{lem*}{$\textbf{Lemma}$}
\newtheorem{pf*}{$\textit{Proof}$}
\newtheorem{prop*}{\textbf{Proposition}}
\newtheorem{cor*}{{\textbf{Corollary}}}
\newtheorem{rem*}{{\bfs \text{Remark}}}
\journal{Pattern Recognition}

\newcommand{\dobib}{ 
    \bibliographystyle{model5-names}\biboptions{authoryear}
    \bibliography{mybib} 
}

\usepackage{newfloat}
\DeclareFloatingEnvironment[
    fileext=los,
    listname={List of Algorithms},
    name=Algorithms,
    placement=tbhp,
    within=none
    ]{algo}
    
\newcommand{\prooflocation}{supplementary material}
\newcommand{\tabledimension}{0.65\linewidth}

\begin{document}

\renewcommand{\dobib}{}

\begin{frontmatter}



\title{Stochastic mean-shift clustering}


\author[1,2]{Itshak Lapidot}
\ead{itshakl@afeka.ac.il}
\address[1]{Afeka Tel-Aviv Academic College of Engineering, School of Electrical Engineering, Israel}
\address[2]{Avignon University, LIA, France}

\author[3]{Yann Sepulcre}
\ead{yanns@mail.sapir.ac.il}
\address[3]{Sapir Academic College, Department of Engineering, Sderot, Israel}

\author[4]{Tom Trigano}
\ead{thomast@sce.ac.il}
\address[4]{Shamoon College of Engineering, Department of Electrical Engineering, Ashdod, Israel}
  
\begin{abstract}

We present a stochastic version of the mean-shift clustering algorithm.  In this stochastic version a randomly chosen sequence of data points move according to partial gradient ascent steps of the objective function. Theoretical results illustrating the convergence of the proposed approach, and its relative performances is evaluated on synthesized 2-dimensional samples generated by a Gaussian mixture distribution and compared with state-of-the-art methods. It can be observed that in most cases the stochastic mean-shift clustering outperforms the standard mean-shift. We also illustrate as a practical application the use of the presented method for speaker clustering.

\end{abstract}


\end{frontmatter}


\section{Introduction}
\label{sec:Introduction}
Clustering algorithms are still actively investigated due to their many areas of application. Numerous algorithms have been proposed and investigated, among which the $k-$means~\citep{lloyd_least_1982}, Spectral clustering~\citep{shi_normalized_2000,ng_spectral_2002}, DB-SCAN~\citep{ester_density_1996}, and the well-known \textit{Mean-shift} (\mshift) clustering algorithm. \mshift\ is an effective non-parametric iterative algorithm~\citep{Fukunaga1975}, which is versatile for clustering, tracking, and smoothing tasks. 
A well-known and used variant of \mshift is the
\textit{blurring mean-shift} (\bms)~\citep{Cheng_1995}. Both \mshift
and \bms algorithms can be coined ``deterministic'' iterative procedures aiming to find local maximizers of an objective function, since they do not involve any random selection of points to perform their update rule. 

Both \mshift and \bms algorithms have been applied to a variety of
domains, and several variations around their original formulation have
been proposed: see~\citep{Carreira-Perpin2006} for \bms with a
Gaussian kernel (known as Gaussian blurring mean-shift); for \bms
applied to high-dimensional data clustering
see~\citep{Chakraborty2021}. Applications in image processing include
image segmentation~\citep{Lu2011}, 
medical and satellite images feature extraction~\citep{Ai2014, Wu2015} and videos segmentation~\citep{Wang2004}. In speech processing, an adaptive version of \mshift has been used to segment speaker clusters \citep{Salmun2016a,Salmun2016b,Salmun2017,Cohen2021}. The mathematical properties of the convergence of these algorithms have also been thoroughly investigated: \citep{Chen_2015,Yamasaki_2024} investigated convergence properties, while \cite{deutsch_weight_2018} provided a unifying view for density-based clustering and clarifies mean-shift-style mode-seeking mechanics for image analysis.

With the growing size of modern datasets and increasing resolution of
data collected by instruments, there is a real need to improve
existing algorithms. One possible direction is to enhance its
convergence rate. Regarding the \mshift for instance, a version involving randomly chosen subsets of points has been proposed in~\citep{Senoussaoui2014}, which is known to be suboptimal; in settings with a large number of data points but low dimensionality, the MeanShift++ (or Extremely Fast Mode-Seeking) proposed in~\citep{Jang_2021} is significantly faster than \mshift with similar performances; subsequent improvement of this method, particularly suitable for high dimensionality datasets can be found in~\citep{Kumar_2023}. Another direction for \mshift improvement is to obtain a better convergence with more significant clusters. \citep{cariou_novel_2022} suggested a robust version of \mshift, which leads to fewer outlier clusters and faster convergence. \cite{yang_mean_2021} proposed a Mean‐Shift Outlier Detector that uses mean-shift density/location updates to reduce bias from outliers and to filter less relevant data points, while \cite{cheng_GBMOD_2025} adapted mean-shift ideas to a granular-ball (coarse-to-fine) representation to improve outlier detection and computational behavior. To address specific structure clustering, \citep{wang_manifold_2010} extended the \bms to a version that could be applied to data lying on a manifold.

In this work we present a stochastic variant of \bms called \textit{\algName} (\shortname), where the data sample points (defining the current state) move one after the other along a climbing path of an objective function, one single point being randomly chosen and moved to a different location at each step; it can still be considered to be a blurring process since the updated points are never replaced to their original locations and keep moving from their current position; at each step any sample point can be picked up for update. 
The paper also presents theoretical results on its convergence.

Random, asynchronous updates in \shortname are preferred for two main reasons. First, updating points one at a time improves convergence stability by preventing the ordinary mean shift from converging to "false modes" that arise when neighboring points use outdated information in noisy or high-dimensional settings. Second, these asynchronous updates allow for in-place computation, which reduces memory usage—a crucial benefit when working with large datasets.

The paper is organized as follows. Section \ref{sec:MeanShiftAlgorithms} presents the main competing clustering algorithms, including the \shortname. We then introduce the theoretical results on \shortname, and discuss its convergence. We present in section \ref{sec:ExperimentsResults} clustering evaluation criteria used for evaluation, and detail results obtained for these clustering algorithms on synthetic datasets. As an application, we also illustrate the application of \shortname on speaker clustering tasks in section~\ref{sec:speech}. 
Proofs of the presented results are detailed in the \prooflocation.

\dobib


\section{Description of the Stochastic Mean Shift (SMS)}
\label{sec:MeanShiftAlgorithms}

The following section presents our stochastic version of \mshift. For comparison, we first recall the Mean-shift (MS) and the Blurring mean-shift (BMS). Table~\ref{tab:notations} summarizes the notations used in this paper.

  \begin{table*}[ht]
    \centering
    \caption{Summary of the main notations used throughout the paper}
    \maxsizebox{0.98\linewidth}{!}{%
      \begin{tabular}{ccp{0.95\linewidth}}
    \toprule
    Notation & \phantom{abcde} & Significance \\
    \midrule
    \phantom{aaaaaaa}         & General notations & \\
    $[n]$ & & the set of integers $\{1 , \ldots , n \} $ \\
    $\re_{+}$ & & the set of non negative reals $[ 0,\infty )$ \\
    $2^S $ & & the collection of all subsets of a set $S$ \\
    $\| \mathbf{u} \| \, , \ \mathbf{u} \in \re^{n\times 1} $ & & the $\ell_2 $ norm
    of $\mathbf{u}$ \\
    \midrule
    \phantom{aaaaaaa}             & Specific notations & \\
    $x_1,\ldots x_J$ & & The $J$ data points \\
    $K(\mathbf{u}) = k\lpar  \| \mathbf{u} \|^2 \rpar \ ; \ G(\mathbf{u}) = - k^{\prime}\lpar\| \mathbf{u}\|^2 \rpar $ & & density kernel $K$ on $\re^n $ associated to the profile $k:[0,\infty) \to \re_{+}$ \, ; \, "weight" function on $\re^n $ \\
    $\neighb{x}, \, \, x \in \re^d$ & & subset of $[n]$ of the neighboring indices $\left\{ {i \, :\left\| {x  - {\point_i}} \right\| < h} \right\}$ \\
    $\ridx{k}$ & & the index $i \in [n]$ randomly chosen at the $k^{th}$ step
    of \shortname \\
    $\nabla_{i} L$ & & partial gradient $\nabla_{\point_i } L $ of $L$ w.r.t the variable $\point_i $ \\
    $\nabla^{(k)} \ ; \ \nabla_{i}^{(k)} $ & &  gradient of $L$ at step $k$ of \shortname \, ; \,  partial gradient $\nabla_{i} L$ at step
    $k$ of \shortname \\
    $\| \mathbf{U} \| \, , \ \mathbf{U} = [\mathbf{u}_{1}^{T} , \ldots , 
    \mathbf{u}_{n}^{T} ]^{T} \, , \, \, \mathbf{u}_i \in \re^{d\times 1} $ & & 
    $ \max\limits_{i \in [n]} \| \mathbf{u}_i \|_2 $: maximal component wise $\ell_2 $ norm value \\
     \bottomrule
  \end{tabular}}
  \label{tab:notations}
\end{table*}


First, we provide details on the specific kernels to be considered in this work. These assumptions will be further used in the theoretical proofs.

\begin{def*}
    \begin{enumerate}
        \item We call a profile function a mapping $k: $[0,1]$ \to \re_{+} $ that is $C^1 $, non-increasing, convex and such that $k^{\prime}(t)<0 $ for all $t \in [0,1)$.
  \item a multivariate function which can be written as $K( \mathbf{u} ) = k\lpar  \| \mathbf{u} \|^2 \rpar$ for all $\mathbf{u} \in \re^d$ 
  will be referred to as a kernel with profile function $k$.
    \end{enumerate}
\end{def*}
Typical examples of such profiles and associated kernels include $k(t)= \lpar 1-t \rpar_{+}^{\alpha}$ for $\alpha=2,3,4$, which lead respectively to the Bi-, Tri-, and Quadweights kernels
 $ K( \mathbf{u} ) = \lpar 1- \| \mathbf{u} \|^2  \rpar_{+}^{\alpha} $, respectively. 

For any sample set $\{ \point_i \}_{i \in [n]}$ with $\point_i \in \re^{d}, i=1\ldots n $, stemming from some unknown multi-modal density $\truedensity$ defined on $\re^d $, given a kernel $K$ and a positive bandwidth parameter $h>0$, the kernel density estimate (KDE) of $\truedensity$ associated with $\{ \point_i \}_{i \in [n]}$ is defined as
 \begin{align}\label{eq:KDE_def}
  \forall \, x \in \re^d \ \,    \kde (x) & = 
  \frac{1}{n h^d }\sum\limits_{i \in[n]} \, K\left ( \frac{1}{h}\, (x - \point_{i} )\right ).
 \end{align}
 Since all the considered kernels have compact support by definition of the profile functions, this property is shared by any KDE defined in \eqref{eq:KDE_def}. For the rest of this paper, 
 we shall refer to the element $\state=[\point_{1}^T, \ldots , \point_{n}^{T} ]^{T} \in \re^{d \, n}$ as the \textit{state} defined by this sample, and interpret any iterative clustering algorithm as a chain of transformations of $\re^{d\,n}$ applied to the initial state $\state^{(0)}$ given by the raw sample until we reach convergence (in a sense to be precised). A state shall be introduced further shortly as 
 $\state = [\point_{i}]_{i \in [n]}$ for the sake of simplicity, without emphasizing further the implicit vector structure.
 
\subsection{Reminders on \mshift and \bms}

\mshift works under the mild assumption that dense regions in the data space correspond to modes of $\truedensity$, which can be estimated by the local maxima of \eqref{eq:KDE_def}. All points are updated to the weighted average of neighboring points, where weights are determined by their distance, until convergence around modes of the estimated distribution
\eqref{eq:KDE_def}.

Mathematically, \mshift is a gradient ascent algorithm on $\kde$ which aims to provide convergence to a local maximizer. This formulation can be found in several references, such as \citep{Senoussaoui2014} and \citep{Comaniciu2002}. For any state $\state$, we shall consider the {\it{mean-shift operator}} $\msop_{h} : \mathbb{R}^d \to \mathbb{R}^d $ which maps a point $x \in \mathbb{R}^d $ to its new position as
 \begin{equation}\label{eq:meanshiftOperator}
 \msop_{h} (x ; \state) = \frac{\displaystyle \sum\limits_{i \in [n]} G\left ( \frac{x - \point_i }{h}\right )\, \point_i  }{\displaystyle \sum\limits_{i \in [n]} G\left ( \frac{x - \point_i }{h}\right )} \ \ \, , \ \ \, 
 \end{equation}
where $G(\mathbf{u}) = - k^{\prime}\lpar\| \mathbf{u}\|^2 \rpar$ defines the non-negative {\it{weight function}} associated with $K$.
Note that the denominator in \eqref{eq:meanshiftOperator} is always positive since $-k^{\prime}(0)>0$. Note also that \eqref{eq:meanshiftOperator} is effectively computed on the {\textit{$\state-$neighborhood}} of $x$ which is defined as $$\neighb{x} = \lset i \in [n] \, | \, G\lpar \frac{x - \point_{i} }{h}\rpar \neq 0 \rset =  \lset i \in [n] \, | \, \| x - \point_{i} \| < h \rset, $$  since $K$ has a radial symmetry.
 The \textit{shift vector} of $x$ is defined as
 \begin{equation}\label{eq:shift_definition}
     \shv (x ; \state) = \msop_{h} (x ; \state) - x \ \ ;
 \end{equation}
 it can be easily verified that $ \shv (x ) $ is indeed proportional to the gradient of 
 $\kde$ taken at $x$, thus allowing to consider \mshift as a gradient ascent method on $\kde$ with variable step-size. As a particular case, when using the Epanechnikov kernel, 
 the shift vector ${m_h}(x)$ is defined as follows:
\begin{equation}
\label{MeanShiftVectorFinal}
{m_h}(x) = \frac{{\sum\limits_{i \in \neighb{x}} {{\point_i}} }}{\# \neighb{x}} - x
\end{equation}
where $\#$ denotes the cardinality of the set ${N_{h}}(x ) $. 
Details on its convergence can be found in~\citep{Yamasaki_2023}.

  


The Blurring mean-shift (\bms) is a related kernel-based iterative method for data clustering, which aims to improve \mshift. Starting from the initial state $\state^{(0)} \in \re^{d n}$ defined by the original data sample, for any $k \in \mathbb{N}_0 $ the $k^{th}$ step of the \bms consists in updating the current state $\state^{(k)}$ to the new state $\state^{(k+1)}$ by applying the mean-shift operator defined as follows:
 \begin{align}\label{eq:kth_step_bms}
     \forall \, i \in [n] \ \ \ & \point_{i}^{(k+1)} = \frac{\sum\limits_{j=1}^n G\lpar \displaystyle \frac{\point^{(k)}_i - \point^{(k)}_j  }{h }\rpar \, \point^{(k)}_j }{\sum\limits_{j=1}^n G\lpar \displaystyle \frac{\point^{(k)}_i - \point^{(k)}_j  }{h} \rpar }
 \end{align}
 In other words, we apply to each sample point the mean-shift operator to find its new location, but by plugging in \eqref{eq:meanshiftOperator} the current state $\state^{(k)}$. Hence, in contrast with \mshift, the points move altogether during the whole process, as summarized in Algorithm~\ref{alg:bms}.


 From a mathematical point of view, the BMS algorithm can be seen as an iterative procedure aimed at converging to a local maximizer of the following non-negative objective function
\begin{align}\label{def:L_function}
    \forall \, \state=[\point_{1}^T , \ldots , \point_{n}^{T} ]^{T} \in \re^{d\,n} \ \ \, L(\state) & = \sum\limits_{1 \leq i \leq j \leq n } K\lpar \frac{1}{h} (\point_i - \point_j )\rpar  \ \ \ ;
\end{align}
Indeed one can reformulate \eqref{eq:kth_step_bms} as a "weighted" gradient step as follows: 
 \begin{align}\label{eq:step_bms_with_gradient}
      \state^{(k+1)} = \state^{(k)}  + \frac{h^{2}}{2} \sum\limits_{i=1}^n \frac{1}{\sum\limits_{j=1}^n G\lpar \displaystyle\frac{\point_{i}^{(k)} - \point_{j}^{(k)}}{h}\rpar} \nabla_{\point_{i}} L (\state^{(k)})
 \end{align}
 where $\nabla_{\point_{i}} L (\state)$ denotes the  gradient component of $L$ related to the $\point_{i}$ direction.
 Any state $\state \in \re^{d n} $ satisfies $\nabla L (\state) = \mathbf{0}_{d n} $ if and only if it is a fixed point of the transformation \eqref{eq:kth_step_bms}, i.e.
\begin{align}\label{eq:critical_point}
    \forall \, i\in [n]  \, \ \ \point_i = \frac{\sum\limits_{j=1}^n G\lpar \displaystyle \frac{\point_i - \point_j  }{h }\rpar \, \point_j }{\sum\limits_{j=1}^n G\lpar \displaystyle \frac{\point_i - \point_j  }{h} \rpar }
\end{align}
  The relation between the stationary points of $L$ and our clustering objective is made explicit in a theorem from~\citep{Yamasaki_2024}, which can be written as follows:
  \begin{theorem*}\label{thm:critical_point_charact}
      Assume the profile function $k$ of the kernel $K$ satisfies assumptions $a)-c)$. A state $\state =[\point_{1}^T , \ldots , \point_{n}^{T} ]^{T} \in \re^{d \, n} $ satisfies $\nabla L (\state)= \mathbf{0}_{d n}$ iff the following is satisfied:     \begin{align}\label{eq:criticalpoint_geometric_char_UPDATED}
          \forall \, (i,j) \in [n]^2 \ \, j \in \neighb{\point_{i}} \, \iff \, \point_j = \point_i 
      \end{align}
  \end{theorem*}
  In other words, for the specific truncated kernels considered in this study, a critical point of $L$ is a state satisfying an ideal clustering condition: two points either coincide, or are distant from $h$ at least. In the previous theorem, \eqref{eq:criticalpoint_geometric_char_UPDATED} will be later used to prove theoretical results on the effectiveness of the stochastic mean-shift, which is introduced in the next subsection. 
 

Convergence of the \bms sequence has been thoroughly investigated and proved~\citep{Chen_2015}; theorems establishing convergence rates for different kernel types can be found in~\citep{Yamasaki_2024}. Though \bms may converge faster than \mshift, its algorithmic complexity is roughly $\mathrm{O}(n^2) $, which makes it computationally expensive for large datasets.


\subsection{Stochastic mean-shift algorithm}
\label{subsec:stochsticMS}
We now present the \textit{stochastic mean-shift} algorithm
(\shortname), which is the main novelty of this paper. Roughly
speaking, \shortname can be seen as a trade-off between \mshift and
\bms, adding some randomness in the shifting process. Its principle is
as follows: starting from an original state ${{\cal X}}^{(0)} = \lbr
{{\point_j}} \rbr_{j = 1}^{n}$, in each subsequent step $k$ an index $i
\in [n]$ is picked up randomly (according to the uniform distribution
and independently of all previous choices), and the chosen point alone is moved by using the shift operator $\msop_{h}$, by plugging in \eqref{eq:meanshiftOperator} the sample $\lbr
\point_{j}^{(k)} \rbr$ considered in its \textit{current state}. In other words, and similarly to \bms, all the sample points may move during the whole process.
 A summarized version of \shortname is described in Algorithm \ref{alg:stochasticMeanShiftClustering}.


\begin{algo*}
  \centering
  \begin{subfigure}{0.3\linewidth}
\scriptsize
\begin{algorithmic}
\For{$\mathbf{x}_i \in \mathcal{X}$}
\While{Convergence is not attained} 
\State $\mathbf{x}_i^{(k+1)} \gets \mathcal{S}_h (\mathbf{x}_i^{(k)};\mathcal{X})$
    \State $k+1 \gets k$
    \State Convergence diagnosis
\EndWhile
\EndFor
\end{algorithmic}
\caption{Mean-Shift}\label{alg:MeanShiftClustering}
  \end{subfigure}
  \begin{subfigure}{0.35\linewidth}
\scriptsize
\begin{algorithmic}
\While{Convergence is not attained}
\For{$\mathbf{x}_i^{(k)} \in \mathcal{X}^{(k)}$}
\State $\mathbf{x}_i^{(k+1)} \gets \mathcal{S}_h (\mathbf{x}_i^{(k)};\mathcal{X}^{(k)})$
    \EndFor
    \State $k+1 \gets k$
    \State Convergence diagnosis
\EndWhile
\end{algorithmic}
\caption{Blurring Mean-Shift}\label{alg:bms}
    \end{subfigure}
  \begin{subfigure}{0.3\linewidth}
\scriptsize
  \begin{algorithmic}
\While{Convergence is not attained}
\State Draw the index $I_k \in [n]$ uniformly

\State $\mathbf{x}_{I_k}^{(k+1)} \gets \mathcal{S}_h (\mathbf{x}_{I_k}^{(k)};\mathcal{X}^{(k)})$
    \State $k+1 \gets k$
    \State Convergence diagnosis
\EndWhile
\end{algorithmic}
\caption{Stochastic
  Mean-Shift}\label{alg:stochasticMeanShiftClustering}
  \end{subfigure}
  \caption{Comparison between Mean-Shift, Blurring Mean-Shift and
    Stochastic Mean-Shift. In all cases, convergence diagnosis is
    either based on a maximal number of iterations or when
    $\|\mathbf{x}_i^{(k+1)}-\mathbf{x}_i^{(k)}\|$ is smaller than a
    user-defined threshold}
  \label{tab:alg_comp}
\end{algo*}



\shortname can be interpreted as some "partial" gradient algorithm: starting from some initial state $\state^{(0)} \in \re^{nd}$, the $k^{th}$ step of \shortname consists in picking randomly an index $\ridx{k} \in \{1,\ldots,n\}$ and defining the $k+1$ state
  as 
  \begin{align}\label{eq:step_sbms_intro}
      \state^{(k+1)} = \state^{(k)} + \frac{h^{2} /2}{\sum\limits_{j=1}^n G\lpar \frac{\point_{\ridx{k}}^{(k)} - \point_{j}^{(k)}}{h}\rpar} \nabla_{\point_{\ridx{k}}} L (\state^{(k)})
  \end{align}
  where $\nabla_{\point_{i}} L (\state)$ denotes the partial gradient of $L$ in the $\point_{i}$ direction. 
It follows immediately that algorithm \ref{alg:stochasticMeanShiftClustering} has a linear complexity $O(k)$, $k$ denoting the overall number of draws. 


\section{Theoretical results}
\label{sec:Theor_results}

We now show that 
 for an \shortname sequence, the points partition into clusters becomes fixed almost surely after a finite number of steps provided one use an appropriate distance threshold to assess similarity (see theorem \ref{thm:clusters_existence}).
We also establish in theorem \ref{thm:convergence_for_one_cluster} 
 that under the restrictive assumption that the distances between any two points in $\state^{(0)}$ are less than $h$, the sample points converge almost surely to the same limit point as $k$ tends to infinity. Though we could not establish in general an a.s. convergence of $\state^{(k)} $, our primary purpose in this work remains effective data clustering, which is indeed achieved by \shortname.

Unless stated otherwise, in this section we suppose given any kernel $K$ defined on $\re^d $, associated to a profile $k$ satisfying all of the assumptions $a)-c)$ exposed in the beginning of section \ref{sec:MeanShiftAlgorithms};  all notations, definitions, and assumptions introduced there will be used in the present section. The presented results' proofs are postponed to the \prooflocation.

\subsection{Ascending property of \algName}
\label{subsec:theory_1}

We consider a state $\state \in \re^{n \,d}$, i.e. an ordered set of $n$ points $ \state= [\point_{1}^T , \ldots , \point_{n}^{T} ]^{T} $ such that $\point_{i} \in \re^d $ for each $i$ ; given also some positive parameter $h > 0 $, we already saw that \shortname can be interpreted as some partial gradient steps process applied to the function
$L(\state) = \sum\limits_{1 \leq i \leq j \leq n } K\lpar \frac{1}{h} (\point_i - \point_j )\rpar $;
 clearly $L$ is non negative, $C^{1}$ and bounded from above by the absolute maximal value $\frac{n(n+1)}{2} k (0)$ which is attained in states such that $\point_i = \point_j  $ for any $1 \leq i,j \leq n$; there are infinitely many critical points of $L$ (in the sense that $\nabla L (\state) = \mathbf{0}_{d n} $) since 
 $L$ is invariant through any isometry of $\re^d $ applied component-wise to $\state$  (these are the states satisfying \eqref{eq:critical_point} as previously mentioned).
Starting from some initial state $\state^{(0)} \in \re^{nd}$, the $k^{th}$ step of \shortname consists in picking randomly an index $\ridx{k} \in \{1,\ldots,n\}$ and defining the $k+1$ state as 
  \begin{align}\label{eq:step_sbms}
      \state^{(k+1)} = \state^{(k)} + \frac{h^{2} /2}{\sum\limits_{j=1}^n G\lpar \frac{\point_{\ridx{k}}^{(k)} - \point_{j}^{(k)}}{h}\rpar} \nabla_{\point_{\ridx{k}}} L (\state^{(k)}),
  \end{align}
 where $\nabla_{\point_{i}} L (\state)$ denotes the gradient of $L$ in the $\point_{i}$ direction. Since the initial state $\state^{(0)}$ is given, at any step $k\geq 1$ of \shortname the state $\state^{(k)}$ depends solely on the random sequence of indices $[\ridx{k-1},\ldots,\ridx{0}]$, i.e. there exists a non-random function $F^{(k)}: [n]^{k} \to \re^{d n}$ such that $\state^{(k)} = F^{(k)} [\ridx{k-1},\ldots,\ridx{0}]$. 
 

In the following, we shall denote the partial gradient $\nabla_{\point_i } L $ w.r.t. the variable $\point_i $ by $\nabla_{i} L$ for simplicity. 

The following preliminary result shows that the value of $L$ is non-decreasing at any step \eqref{eq:step_sbms} of \shortname.
  \begin{prop*}\label{prop:non_decreasing_L}
      Given any initial state $\state^{(0)} \in \re^{d n}$, for any $k \in \mathbb{N}$ it holds that 
      \begin{align}\label{eq:non_decreasing_L} 
      L\lpar \state^{(k+1)} \rpar - L\lpar \state^{(k)} \rpar \geq C \, \| \point_{\ridx{k}}^{(k+1)} - \point_{\ridx{k}}^{(k)} \|^2  
      \end{align}
      for some constant $C>0$, with $\ridx{k} \in [n]$ denoting the randomly chosen index at step $k$. Two consequences follow from \eqref{eq:non_decreasing_L}: 
      \begin{enumerate}[label=\arabic*$)$]
      \item the sequence $\lpar L(\state^{(k)})\rpar_{k \in \mathbb{N}} $ of the $L$ values of the \shortname states is non decreasing and convergent ; 
      \item it holds furthermore that for a positive constant $D$ (depending only on $n$, $h$ and the kernel $K$):
 \begin{align}\label{eq:bounded_partial_gradient}
     \| \nabla_{\ridx{k}} L (\state^{(k)}) \| \leq D \,\sqrt{ L\lpar \state^{(k+1)} \rpar - L\lpar \state^{(k)} \rpar } \ ; 
 \end{align}
 hence $ \nabla_{\ridx{k}} L (\state^{(k)}) \to \mathbf{0}_{d} $ a.s as $k \to \infty$. 
      \end{enumerate}
  \end{prop*}

  \begin{rem*}
      An easy consequence of \eqref{eq:bounded_partial_gradient} is that for any $\epsilon >0$, it holds a.s that
      \begin{align} 
      \# \lset j \, | \, \ \| \nabla_{\ridx{j}} L (\state^{(j)}) \| \geq \epsilon \rset \leq \lpar \frac{D}{\epsilon} \rpar^2 \lbr \frac{n(n+1)}{2}k(0) - L(\state^{(0)}) \rbr \nonumber
      \end{align}
  \end{rem*}

  \subsection{Clustering and convergence results for \shortname}
\label{subsec:theory_2}
Proposition \ref{prop:non_decreasing_L} showed that in the \shortname algorithm, the sequence of partial gradients of $L$ (computed at each step w.r.t. the randomly selected variable) tends to $0$ a.s. Since for each $k \in \mathbb{N}$ the index $\ridx{k}$ is drawn from the uniform distribution on $[n]$ and independently of the sequence $\lbr \ridx{k-1},\ldots, \ridx{1} \rbr $, we will show at proposition \ref{prop:nabla_goes_to_zero_as} that the whole gradient $\nabla L (\state^{(k)}) \to \mathbf{0}$ a.s. 

We first introduce sequences of stopping times, which shall be used in the proofs. In all of the sequel we use the following notations: at step $k \in \mathbb{N}$ of the \shortname algorithm, the gradient $\nabla L (\state^{(k)})$ shall be denoted by $\nabla^{(k)}$; hence $\nabla_{i}^{(k)}$ shall denote $\nabla_{\point_i } L (\state^{(k)})$ for any $i \in [n]$; and $\| \nabla^{(k)} \|$ shall denote the norm $\max\limits_{i \in [n]} \| \nabla_{i}^{(k)} \|_2 $ i.e the maximal component wise $\ell_2$ norm value of the gradient $\nabla^{(k)}$. 
   
 Suppose given now any $\epsilon >0$ and any fixed positive integer $k_0 $ ; for any $t > k_0$ we define the following event $\SmallSubg{t} = \lset \forall \, k \in (k_0 , \ldots , t] \  \| \nabla_{\ridx{k}}^{(k)} \| < \epsilon \rset $ ; in other words $\SmallSubg{t}$ occurs if and only if for each $k_0 <k \leq t $  the partial gradient computed w.r.t the selected variable $\point_{\ridx{k}}$ has a norm smaller than $\epsilon$. We define a first stopping time as 
 \begin{align}\label{def:stopping_1}
     T_1 = \min \lset t>k_0 \, | \ \| \nabla^{(t)} \| \geq \epsilon \, \, \wedge \, \, A_t \rset \ ; 
 \end{align}
 in case there is no $t > k_0$ satisfying the condition above, then we set $T_1 = \infty $. We further define for any integer $p \geq 2 $ a $p^{th}$ stopping time $T_p $ by the following recursive way:
 \begin{align}\label{def:stopping_p}
     T_p = \min \lset t>T_{p-1} \, | \ \ \| \nabla^{(t)} \| \geq \epsilon \, \, \wedge \, \, \SmallSubg{t} \rset \ , 
 \end{align}
 setting $T_p = \infty $ in case $T_{p-1}=\infty $ or if there is no $t>T_{p-1}$ satisfying the condition.
 Note that the sequence $(T_p )_{p \in \nn }$ is indeed a sequence of stopping times w.r.t the \shortname indices sequence $\lbr \ridx{k} \rbr_{k \in \nn}$ since for any $p \in \nn $ and $k \in \nn $ it holds that 
 $\lpar T_p = k \rpar \in \sigma \lset \ridx{j} \, ; \, \, j \in [k] \rset $. 
 
\begin{lem*}\label{lem:preliminary_to_prop_on_stopping_times}
  Given any positive real $\epsilon >0$ and fixed positive integer $k_0 $ as before, define the sequence $(T_p)$ of stopping times as in \eqref{def:stopping_1}\eqref{def:stopping_p}; 
  for any $p \in \nn$ we define the event $ \Omega_p = \{ T_p < \infty \}$. 
  The following conclusion holds:
$\prob{\Omega_{p}} \leq \lpar 1 - \frac{1}{n}\rpar^p $, hence 
  $\lim\limits_{p \to \infty} \prob{\Omega_{p}} = 0$. 
\end{lem*}

Lemma \ref{lem:preliminary_to_prop_on_stopping_times} yields the following proposition.
\begin{prop*}\label{prop:nabla_goes_to_zero_as}
Given any initial state $\state^{(0)} \in \re^{d n}$, the \shortname sequence satisfies $\nabla^{(k)} \to \mathbf{0}_{d n}$ holds a.s as $k \to \infty$. 
\end{prop*}

The main clustering result is now presented. It states that almost surely, after a specific step in the \shortname, well-separated and stable clusters are observed. 

    \begin{theorem*}\label{thm:clusters_existence}
    Given any state $\state^{(0)} \in \re^{d n}$, let $\lpar \state^{(k)} \rpar $ be the \shortname states sequence ; 
        suppose $\tau$ is any given real in $(0,\frac{h}{2})$. 
        Then there exists a.s some partition $J_1 ,\ldots,J_M $ of $[n]$ 
        and $K \in \nn $ such that for every $k \geq K$:
        \begin{enumerate}[label=\arabic*$)$]
            \item for any $l\in [M]$ it holds that 
            $\max\limits_{\substack{i \in J_l \\ j \in J_l }} \| \point^{(k)}_{i} - \point_{j}^{(k)} \| < \tau \ \ ; $ 
            \item for any $(l,m) \in [M]^2 $ s.t $l \neq m$, it holds that
            $ \min\limits_{\substack{i \in J_l \\ j \in J_m }} \| \point^{(k)}_{i} - \point_{j}^{(k)} \| > h- \tau \ \ ; $
        \end{enumerate}
        in other words a.s. the points $\point_{1}^{(k)}, \ldots,\point_{n}^{(k)}$ belong to fixed clusters for sufficiently large $k$, 
        the $l^{th}$ cluster being defined as $\mathcal{C}_{l} = \lset \point_{i}^{(k)} \, , \, i \in J_l \rset $ (for all sufficiently large $k$); moreover all clusters diameters tend a.s to $0$ as $k \to \infty$.  
    \end{theorem*}

  It could be noted that the existence of such fixed clusters does not imply that a.s. the state $\state^{(k)}$ converges in $\re^{d n}$. Though we could not generally prove that the \shortname sequence converges a.s., we establish below a convergence result in the particular case of a sample of diameter less than $h$ (thus yielding a single cluster whichever trajectory the \shortname may take).
  The first part of the proof follows a similar argument from~\citep{Chen_2015}.
  \begin{theorem*}\label{thm:convergence_for_one_cluster}
      Suppose $\left ( \state^{(k)} \right )_{k \in \mathbb{N}_{0}} $ is a \shortname sequence obtained with a kernel $K$ associated to a profile $k$ satisfying assumptions $a)-c)$. Suppose the initial state $\state^{(0)} \in \re^{d n} $ satisfies that 
      $\forall \, (i,j) \in [n]^2 \ \| \point_{i}^{(0)} - \point_{j}^{(0)} \| < h$. Then a.s there exists $\point \in \re^d $ such that $\forall \, i \in [n] \ \lim\limits_{k \to \infty} \point_{i}^{(k)} = \point $. 
  \end{theorem*}

\subsection{Practical Convergence Diagnosis}

Given an initial state $\state^{(0)} \in \re^{d n}$, the minimal number $k$ of \shortname steps which are necessary to achieve the aforementioned clustering is by nature a random variable. No theoretical results in this paper provide further precision about a higher bound on $k$ which would hold under a specified high probability. Though each step costs linearly in $n$ since we shift only one point at a time, an \shortname process as described in algorithm 
 \ref{alg:stochasticMeanShiftClustering} might need more steps than
 its \bms or \mshift counterparts in order to achieve the desired
 clustering. This constraint makes it necessary to define convenient
 stopping criteria, even if the clustering above has not been
 completely achieved yet. In practice, whenever a sufficient number of points have a sufficiently small last shift, the algorithm stops. Such a practice is justified by the conclusions of propositions \ref{prop:non_decreasing_L} and \ref{prop:nabla_goes_to_zero_as}. 


\section{Results and Discussion on Synthetic Data}
\label{sec:ExperimentsResults}

This section presents the results and their discussion for simulated, synthetic data. In synthetic data, we compared the behavior of \mshift, \bms, and \shortname. All the experiments were performed using MATLAB R2025a, on an i9 Intel computer with 32 GB of RAM. Code is made available upon request for reproducibility.

\subsection{External evaluation criteria for clustering performance}
\label{subsec:ClusteringEvaluationCriteria}
Though clustering performance can be assessed using intrinsic, task-independent, criteria (see e.g. \citep[chapter 17]{Manning2008} or~\citep{Bolshakova2003, Tibshirani2001}), 
Instead, we shall follow a second common approach, which typically relies on external, task-dependent evaluation criteria. More specifically, clustering results will be compared to predefined labels describing the data.  
In this framework, a common way to evaluate clustering performance is using a notion of cluster or class purity, which can be expressed in several ways. The general idea is to quantify similarity between two partitions of the data sample, the first induced by the predefined classes and the other induced by the clustering outcome. 
In this paper, we rely on the notions of the quantity $K$ introduced
below in \eqref{def:K}, which is itself based on the notions of
\textit{average cluster purity} ($ACP$) and \textit{average label
  purity} ($ALP$) defined in~\citep{Cohen2021}, and on the quantity
$G$ introduced in \eqref{eq:G_definition}, which itself relies on the
notions of cluster purity. We briefly recall their definitions for convenience.
Following \citep{Manning2008}, our first evaluation criteria are introduced below, and rely on the following two quantities:
\begin{align}\label{EvalCriterionIntersection}
{Pur(\mathcal{C},\mathcal{D})} = \frac{1}{N}\sum\limits_{q = 1}^Q \underset{d_r \in \mathcal{D}}{\max} \, n_{q\,r } \ \ \, ; \ \,  
{Pur(\mathcal{D},\mathcal{C})} = \frac{1}{N}\sum\limits_{r = 1}^R \underset{c_q \in \mathcal{C}}{\max} \, n_{q\,r }  
\end{align}
where $n_{q,r}$ denotes the total number of data points in cluster
$c_q$ associated with the true label $d_r$, $\mathcal{C} = \{c_q \ ; \
1\le q \le Q\}$ is the set of all the obtained clusters and
$\mathcal{D} = \{d_r \ ; \ 1\le r \le R\}$ is the set of all true labels.
$Pur(\mathcal{C},\mathcal{D})$ is traditionally defined
as the clustering "purity" (w.r.t. the ground truth) and is equal to
the optimal success rate that can be achieved by a rule assigning all
points of a given cluster to one class. Similarly, $Pur(\mathcal{D},\mathcal{C})$ defines class purity
 (w.r.t the clustering). Our first measure of clustering performance shall be the following trade-off quantity:
\begin{align}\label{eq:G_definition}
    {G = \sqrt {Pur(\mathcal{C},\mathcal{D}) \cdot
  Pur(\mathcal{D},\mathcal{C})} }  \ \ \in (0,1].
\end{align}

The purpose of the quantities $ACP$ and $ALP$ introduced below is to define how far the probability of belonging to a cluster conditionally on its label and the probability of having a given label conditionally on the belonging cluster differ from Kronecker's delta distributions, in an average sense, and for uniform priors. More specifically, we define
\begin{align}
    ACP & = \frac{1}{Q} \sum\limits_{q=1}^{Q} \sum\limits_{r=1}^{R} P(d_r | c_{q})^2 \label{eq:ACP_def} \\ 
    ALP & = \frac{1}{R} \sum\limits_{r=1}^{R} \sum\limits_{q=1}^{Q} P(c_{q} | d_r )^2  \label{eq:ALP_def} 
\end{align}
where in the latter 
$P(d_r | c_{q})$ the probability for a point of the cluster $c_q$ to
have the true label $d_r$ and $P(c_{q} | d_r )$ the probability for a
point with known label $d_r$ to belong to the cluster $c_q$,
respectively. The inner sums in \eqref{eq:ACP_def} and
\eqref{eq:ALP_def} can also be seen as a purity index of the cluster $c_q$ and a purity index for label $d_r$, respectively. 
Roughly speaking, lower values of $ACP$ occur whenever the clusters
are not class homogeneous. In contrast, lower values of $ALP$ occur whenever the classes are decomposed along several clusters (both in an average sense). Taking both aspects into account can be done by considering the geometric average, which will also be investigated: 
 \begin{align}\label{def:K}
   K = \sqrt {ACP \cdot ALP} \ \ \in (0,1]
\end{align}

\subsection{Multi-modal data clustering}
\label{subsec:MultiModal}

We present clustering results for visual inspection and
discussion. The synthetic data was generated using a bidimensional Gaussian mixture of three Gaussians. The common hyperparameters and the different sizes used for the data clusters are summarized in Table~\ref{tab:multimodal}.

\begin{table}
  \centering
  \caption{Hyperparameters used in multi-modal clustering experiments}
  \maxsizebox{\tabledimension}{!}{%
    \begin{tabular}{lccccc}
    \toprule
& \phantom{abcdefg}    & Set 1 & Set 2 & Set 3 & Set 4 \\
    \midrule
    Number of clusters & & \multicolumn{4}{c}{3} \\
    Gaussian expectations& & \multicolumn{4}{c}{$m_1 =
                            [1\ 1]^T$,
                            $m_2 = [-1\ -1]^T$, $m_3 = [1 \ -1]^T$}
    \\
    Covariance matrix & &\multicolumn{4}{c}{$0.64\, \mathbf{I}_2$} \\
    Cluster 1 size $n_1$ & & 250 & 50 & 1500 & 100 \\
    Cluster 2 size $n_2$ & &  250 & 50 & 1500 & 300 \\
    Cluster 3 size $n_3$ & & 250 & 50 & 1500 & 50 \\
    \bottomrule
  \end{tabular}}
  \label{tab:multimodal}
\end{table}
For all these experiments, we performed 20 runs of the three clustering algorithms presented in the previous section: \mshift, \bms and \shortname. In all the cases, we choose for all the algorithms the inner parameters (radius, cluster merging threshold and so forth) that are the most favorable in average  for \mshift, to make the comparison as fair as possible (and even slightly biased against the proposed approach).

\begin{figure}[ht]
  \centering
  \includegraphics[width=0.95\linewidth]{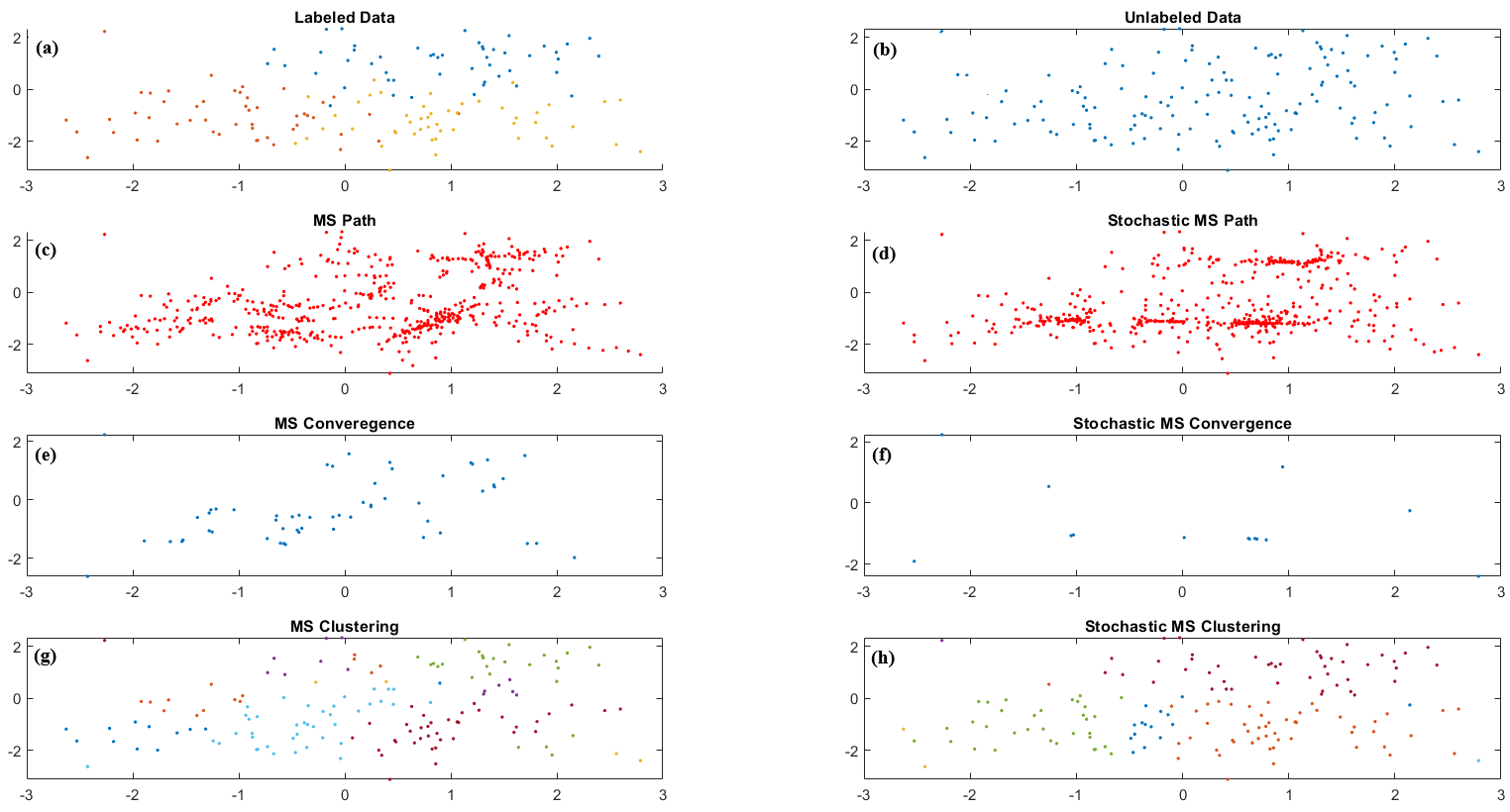}
    \caption{ Results of experiment 2 (a)
      data with true labels; (b) unlabeled data; (c) converging paths
      of \mshift; (d) converging paths of \shortname; (e) the found
      modes of \mshift; (f) the found modes of \shortname; (g) the
      clustering of \mshift; (f) the found clustering of \shortname.}
      \label{fig:ResultsExample}
\end{figure}

Figure \ref{fig:ResultsExample} presents some results of the clustering
process for set 2 with 100 iterations, and compares the obtained
trajectories for \mshift and \shortname.
Note that, for \mshift, the directed paths to the modes can
be observed, while for \shortname (and also for \bms trajectories) they are much less visible. However,
the high-density areas at the modes remain clear. Sub-plots (e) and (f)
shows the found modes of both algorithms, while 
sub-plots (g) and (h) show the final clusters of the
deterministic and the stochastic algorithms, respectively. In this specific example, deterministic mean-shift clustering ended with $14$ clusters while stochastic mean-shift clustering provided only $9$ clusters.

Overall results are displayed in
Table~\ref{tab:global_clustering_results}.
\begin{table}[ht]
  \centering
  \caption{Average of 20 numerical criteria found on experiments 1 to 4. Best results are emphasized.}
  \maxsizebox{\tabledimension}{!}{%
    \begin{tabular}{ccccccccc}
    \toprule
    & &$ACP$&$ALP$&$K$& $Pur(\mathcal{C},\mathcal{D})$&
                                                     $Pur(\mathcal{D},\mathcal{C})$&
                                                                        $G$
      & Number of clusters \\
      \midrule
    \multirow{3}{0.1\linewidth}{\rotatebox{90}{Set 1}}
    &\mshift  & 0.86 & 0.86 & 0.86 & \textbf{0.93} & 0.93 & \textbf{0.93} & 4.2 \\ 
    &\bms & 0.88 & \textbf{0.87} & 0.88 & 0.92 & \textbf{0.93} & 0.92 & 3.5 \\
    &\shortname &  \textbf{0.92} & 0.86 & \textbf{0.89} & 0.93 & 0.93 & 0.93 & 6.2 \\
    \midrule
    \multirow{3}{0.1\linewidth}{\rotatebox{90}{Set 2}}   &\mshift
      &  0.89 & 0.82 & 0.86 & 0.92 & 0.90 & 0.91 & 4.6\\
    &\bms & 0.89 &\textbf{0.86} & \textbf{0.88} & 0.91 & \textbf{0.92} & \textbf{0.92} & 3.7 \\
    &\shortname &  \textbf{0.93} & 0.82 & 0.87 & \textbf{0.93} & 0.90 & 0.901 & 6.3\\
    \midrule
        \multirow{3}{0.1\linewidth}{\rotatebox{90}{Set 3}}   &\mshift
      & 0.88 & 0.87 & 0.87 & 0.93 & \textbf{0.93} & 0.93 & 3.9 \\
    &\bms & 0.89 & \textbf{0.87} & 0.88 & \textbf{0.93} & 0.93 & \textbf{0.93} & 3.8\\
    &\shortname &  \textbf{0.92} & 0.87 & \textbf{0.90} & 0.93 & 0.93 & 0.93 & 7.1 \\
    \midrule
    \multirow{3}{0.1\linewidth}{\rotatebox{90}{Set 4}}   &\mshift
      &  0.86 & 0.83 & 0.85  & 0.92 & \textbf{0.95} & 0.93 & 3.35 \\
    &\bms & 0.88 & \textbf{0.84} & 0.86 & 0.95  & 0.95 & \textbf{0.95} & 3.9 \\
    &\shortname &  \textbf{0.91} & 0.83 & \textbf{0.87} & \textbf{0.95} & 0.94 & 0.94 & 6.4\\
    \bottomrule
  \end{tabular}}
  \label{tab:global_clustering_results}
\end{table}
It can be seen that the stochastic mean-shift performed better than
the deterministic mean-shift in many cases, and exhibits performance
comparable with \bms. 
In most cases, the stochastic mean-shift produced almost twice as many clusters as \mshift. Thus, the ALP in the stochastic case was relatively low. Indeed, the fact that points are drawn one after the other uniformly in such a case tends to leave the outlier points with little change, thus artificially increasing the number of clusters and diminishing \shortname's performance.

\begin{figure}[ht]
\centerline{\includegraphics[width=\columnwidth]{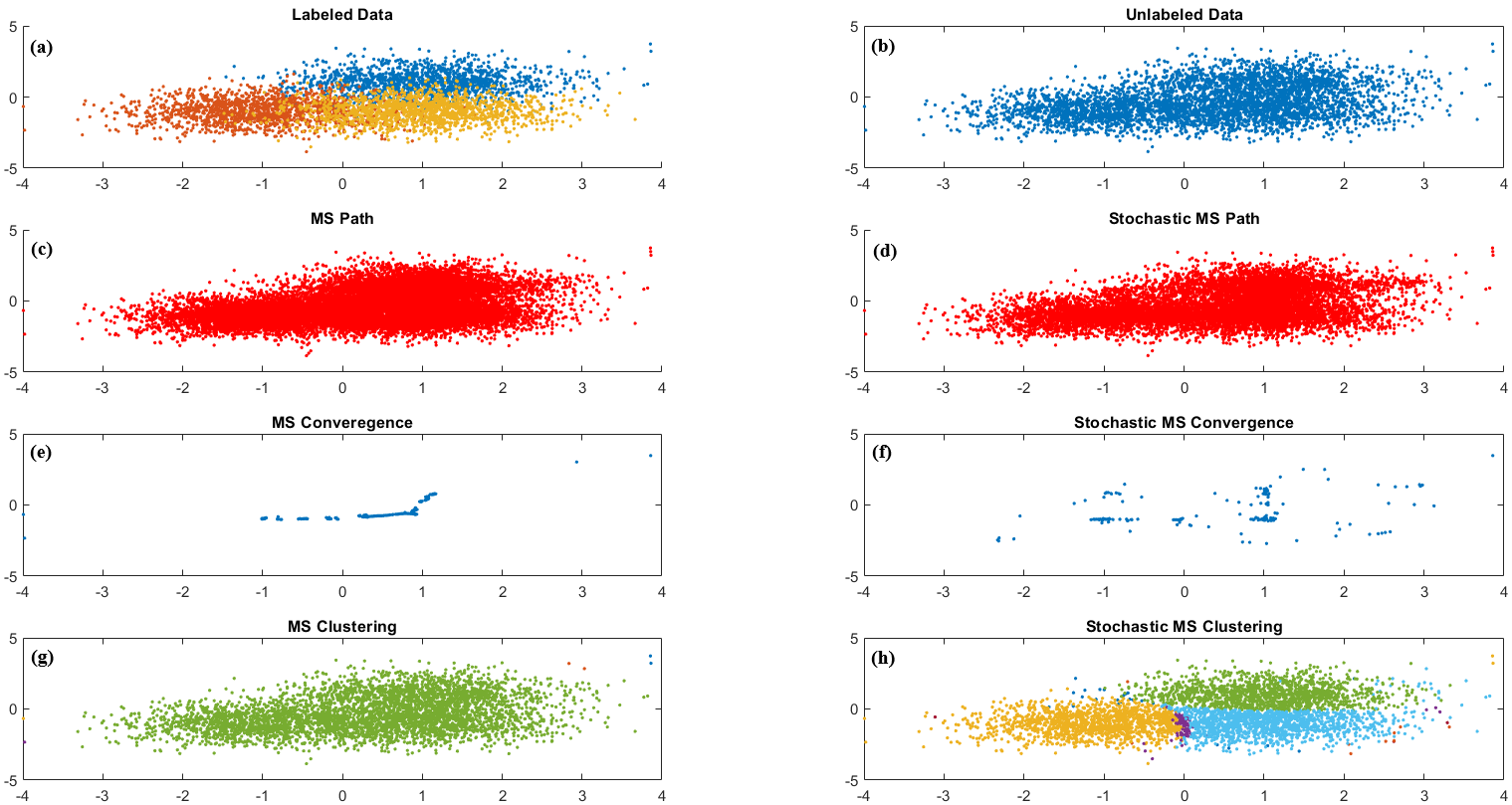}}
    \caption{ Results of Set $4$ (a) data with true labels; (b) unlabeled data; (c) converging paths of the deterministic mean-shift; (d) converging paths of the stochastic mean-shift; (e) the modes found of the deterministic mean-shift; (f) the modes found of the stochastic mean-shift; (g) the clustering of the deterministic mean-shift; (f) the found clustering of the stochastic mean-shift.} 
    \label{fig:ResultsExample1500}
\end{figure}

%
Set 3 has the most data points per class ($1500$). We notice that the
number of clusters for the stochastic mean-shift is higher than for
other cases. For visual inspection, we can rely on Figure
\ref{fig:ResultsExample1500}. In that case, The data is very dense and in the deterministic case, as at the end of the convergence of one sample, it is placed back to the original place before starting with the next datum, all the modes are very close to each other as can be seen in subplot \ref{fig:ResultsExample1500}e. In contrast, in the stochastic case, no data point is placed back and at any step is updated from its current position if picked up (subplot \ref{fig:ResultsExample1500}f). In the deterministic case large clusters are generated during the merging along with several outliers (subplot \ref{fig:ResultsExample1500}g), whereas in the stochastic case we can see three well-separated clusters (subplot \ref{fig:ResultsExample1500}h), along with one small cluster at the intersection of the other three, the rest being outliers. 

Regarding Set 4, the data is unbalanced, but the number of data points
remains reasonable. From our observation, strategies updating the data's position at each iteration (whether deterministically like \bms or randomly like \shortname) performed better. Indeed, updating the data positions at each iteration allows, in some sense, to escape from the attraction of large clusters, which might explain the results obtained.  

\subsection{\shortname Performance Assessments}

We now detail some experiments performed to evaluate several aspects of the proposed approach (each performed 100 times to get more general insight). Table~\ref{tab:params_global_assessment} summarizes global information on the experiments performed.

\begin{table}[!ht]
  \centering
  \caption{Hyperparameters used for the simulations}
\maxsizebox{\tabledimension}{!}{%
  \begin{tabular}{ccccc}
    \toprule
    \multicolumn{5}{c}{Common hyperparameters} \\
    \midrule
Maximal number of iterations  &   \multicolumn{4}{c}{$10^7$} \\
Kernel radius (bandwidth) &  \multicolumn{4}{c}{$1$} \\
Sample cluster distribution & \multicolumn{4}{c}{$\mathcal{N}(\boldsymbol\mu_n,\boldsymbol\Sigma_n), \ \boldsymbol\Sigma_n = 0.6 \mathbf{I}_n$}\\
Metric & \multicolumn{4}{c}{$\displaystyle \|\mathbf{x}\|_2^2 = {\sum_{k=1}^d x_i^2}$}\\
    Kernel type & \multicolumn{4}{c}{Uniform}\\
    Merging cluster threshold & \multicolumn{4}{c}{$1/3$} \\
    Convergence threshold & \multicolumn{4}{c}{$\mathrm{Change\ in\ position} < 10^{-6}$} \\
    \midrule
&    \parbox[c]{0.25\linewidth}{\centering Algorithmic\\ complexity} & \parbox[c]{0.25\linewidth}{\centering Influence of the class imbalance} & \parbox[c]{0.25\linewidth}{\centering Influence of the number of clusters} & \parbox[c]{0.25\linewidth}{\centering Influence of dimensionality} \\
    \midrule
    Number of clusters & 3 & 3 & $\{2 \ldots 20\}$ & 3 \\
  & $[1 \, 1]^T$  & $[1 \, 1]^T$ & & \\
        $\boldsymbol\mu_n$ & $[-1\, -1]^T$  & $[-1\, -1]^T$ & $\sim \mathcal{UD}(-d/2,d/2)^2$ & $\sim \mathcal{UD}(\{-1,1\})^d$ \\
 & $[1\, -1]^T$  & $[1\, -1]^T$ & & \\
    Samples per cluster & $\{10\ldots 10^4\}$  & $\approx 250$ & $\approx 250$ & $\approx 250$ \\
    Samples dimension & 2  & 2 & 2 & $\{2\ldots 12\}$ \\
    Sample ratio per cluster & $\approx 1$ & from $0.1$ to $10$ & $\approx 1$ & $\approx 1$\\
    \bottomrule
  \end{tabular}
}  \label{tab:params_global_assessment}
\end{table}

\subsubsection{Algorithmic Complexity Assessment}

\begin{figure}
  \centering
  \includegraphics[width=0.95\linewidth]{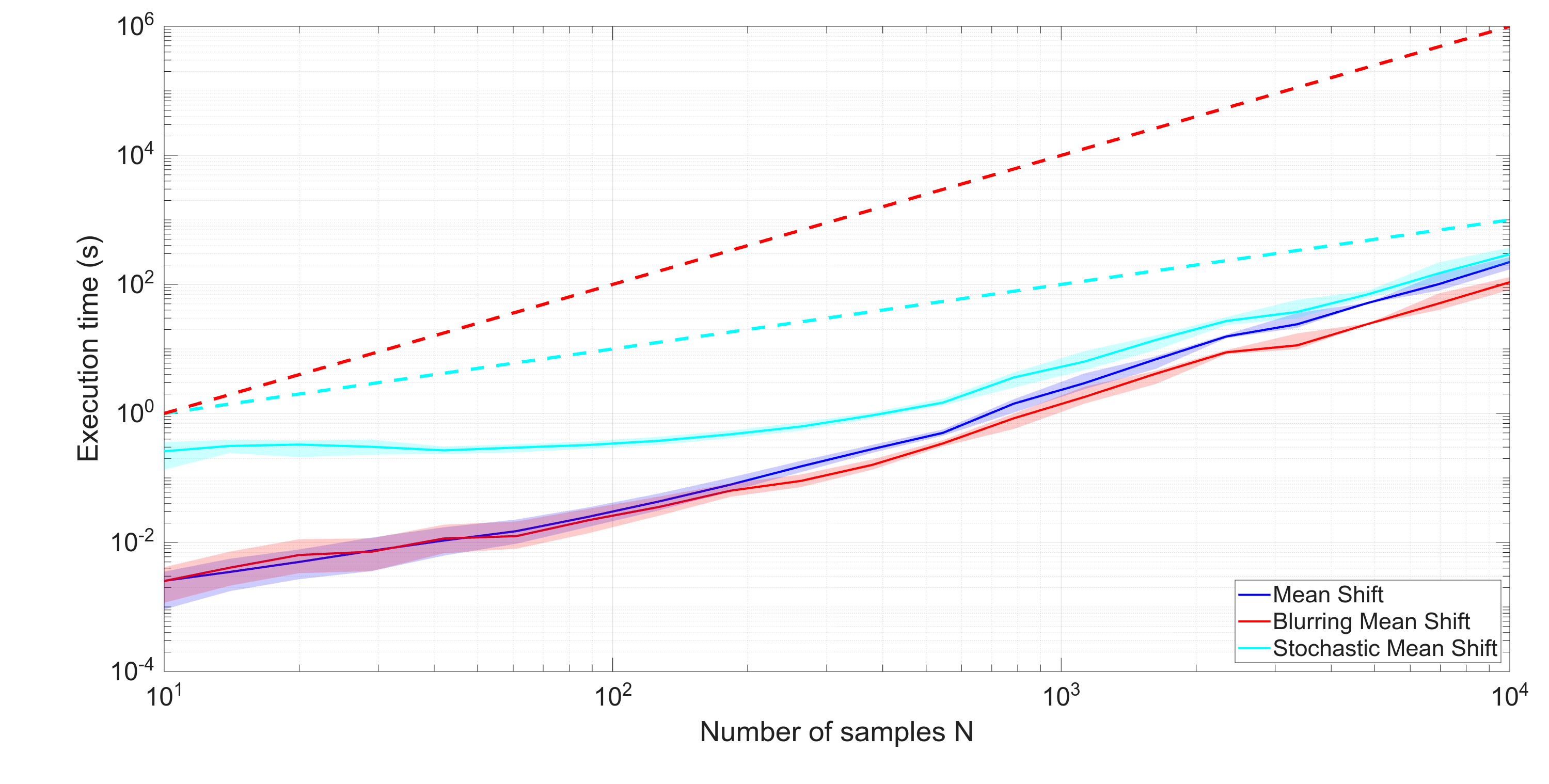}
  \caption{Execution times for \mshift (blue), \bms (red), and
    \shortname (cyan), as a function of the number of points per cluster. Dotted lines illustrate linear complexity (cyan) and quadratic complexity (red) for visual comparison.}
  \label{fig:exec_times}
\end{figure}

The first experiment compared the execution times of the three mean-shift-based clustering methods. The following experiment was repeated 100 times: given a fixed, identical number of samples per cluster, three bidimensional clusters were generated based on three 2D-Gaussian distributions. Clustering was performed using \mshift, \bms and \shortname, and the execution times were saved. The median execution times with associated $90\%$ confidence intervals are presented for all the methods in figure~\ref{fig:exec_times}.

Figure~\ref{fig:exec_times} clearly illustrates the quadratic complexity $\mathrm{O}(N^2)$ of both \mshift and \bms in contrast with the linear complexity $\mathrm{O}(N)$ of the proposed approach. However, this claim must be tempered by the fact that, being stochastic by nature, the proposed algorithm generally requires more iterations to reach convergence than its counterparts. As a result, for problems of modest size, the deterministic approach can be more efficient, since its higher per-iteration cost is offset by faster convergence and the absence of sampling noise. However, as the sample size $N$ increases, the quadratic growth in computational burden of the deterministic method quickly dominates. At the same time, the stochastic algorithm's linear scaling allows it to remain computationally tractable even when many more iterations are needed. We therefore infer that \shortname can become increasingly advantageous for high-dimensional problems, where its lower asymptotic complexity compensates for its slower convergence dynamics.

\subsubsection{Influence of the Cluster Imbalance}

\begin{figure*}[ht]
  \centering
  \begin{subfigure}{0.45\linewidth}
    \centering
  \includegraphics[width=0.9\linewidth]{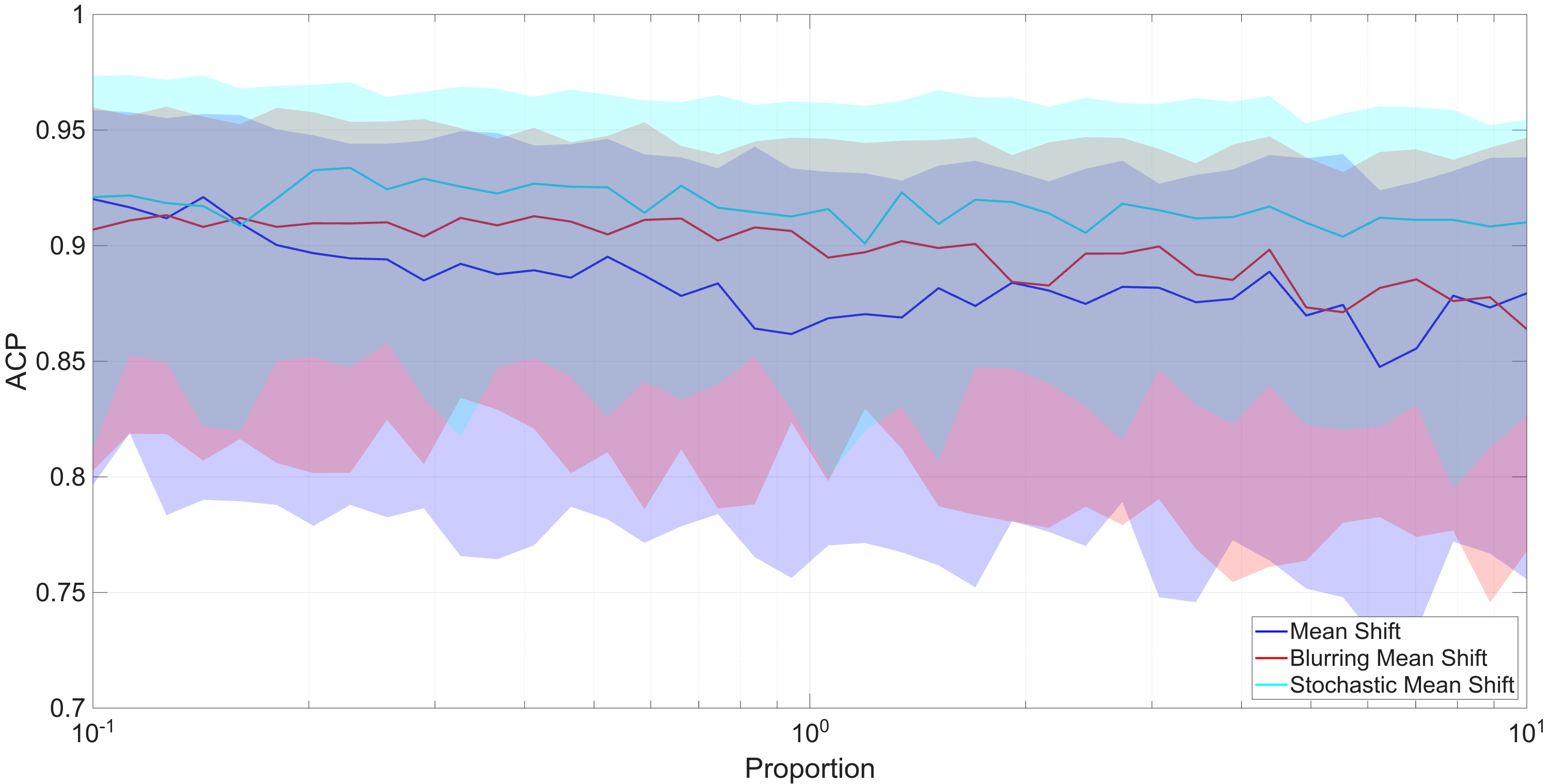}
  \caption{Influence on ACP}
  \label{fig:class_imbalance_ACP}
\end{subfigure}
\begin{subfigure}{0.45\linewidth}
  \centering
  \includegraphics[width=0.9\linewidth]{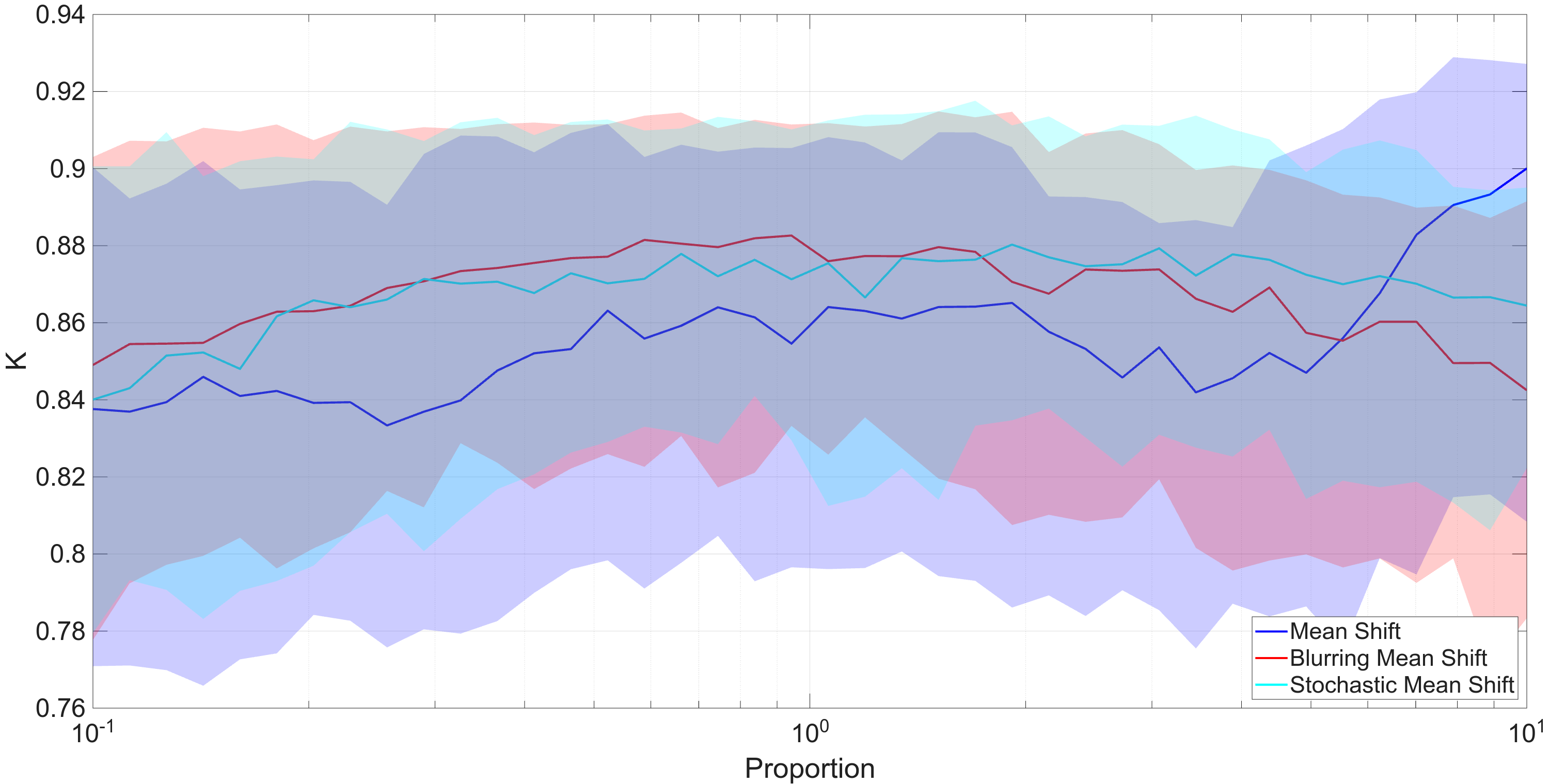}
  \caption{Influence on K.}
  \label{fig:class_imbalance_K}
\end{subfigure}
\caption{Influence of the class imbalance on clustering results for
  \mshift (blue), \bms (red) and \shortname (cyan).}
\label{fig:class_imbalance}
\end{figure*}

We present results for ACP and K when the proportion of points belonging to one specific cluster evolves. In this experiment, we wish to investigate how all algorithms behave when one class becomes significantly dominant with respect to the others. Median results and 90\% confidence intervals are presented in Figures~\ref{fig:class_imbalance_ACP} and \ref{fig:class_imbalance_K} for ACP and K, respectively.

It can be understood from figure~\ref{fig:class_imbalance_ACP} that \shortname generally outperforms \mshift and \bms in terms of average cluster purity. Though the deterministic mean-shift seems to perform better for significant imbalance when looking at K, the observation is misleading. Indeed, when one cluster largely dominates the others, the ALP tends to be one for the ordinary mean-shift, as all the data tend, in that case, to aggregate to a single cluster. This leads to an artificial increase of the K value for large disproportions between clusters, whereas \bms and \shortname performance remain consistent.

\subsubsection{Influence of the Data Dimensionality}

\begin{figure*}[ht]
  \centering
  \begin{subfigure}{0.45\linewidth}
    \centering
    \includegraphics[width=0.9\linewidth]{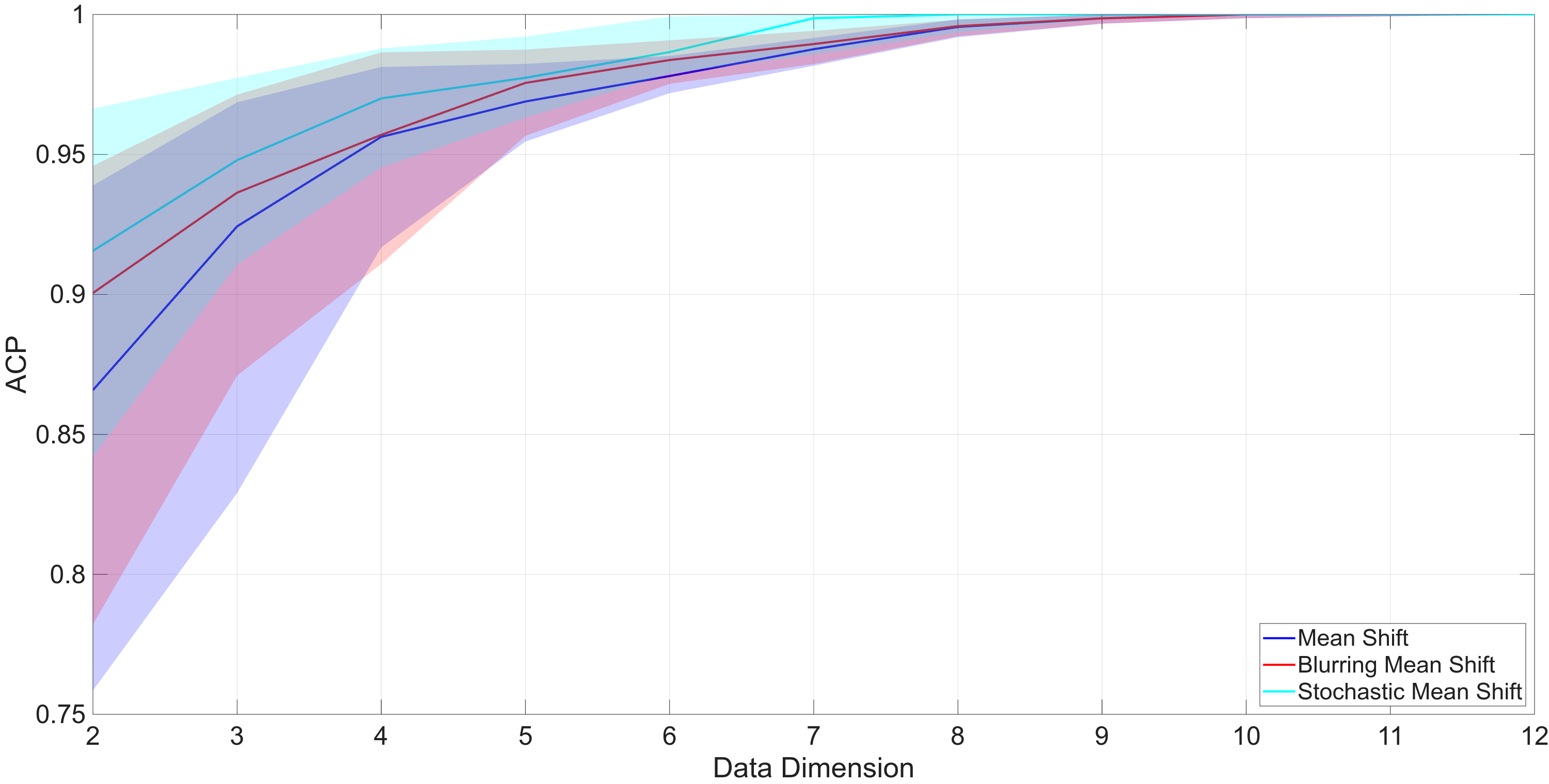}
    \caption{Influence on ACP.}
    \label{fig:dim_ACP}
  \end{subfigure}
  \begin{subfigure}{0.45\linewidth}
  \centering
  \includegraphics[width=0.9\linewidth]{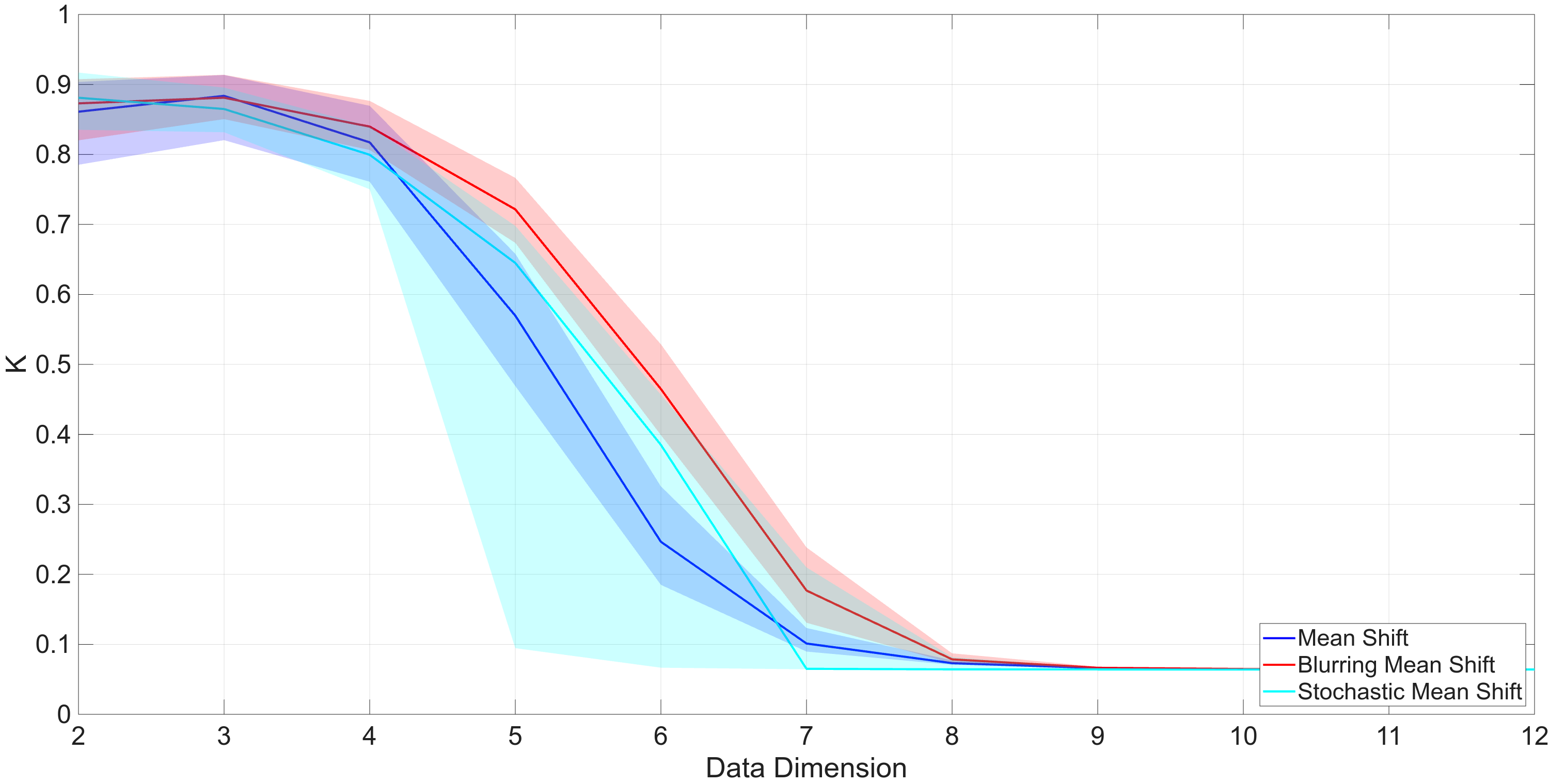}
  \caption{Influence on K.}
  \label{fig:dim_K}  
  \end{subfigure}
  \caption{Influence of the data dimension on clustering results for
    \mshift (blue), \bms (red) and \shortname (cyan).}
  \label{fig:dim}
\end{figure*}

Besides the number of points per cluster, their dimension also strongly influences the clustering results. For this experiment, the Gaussian averages were drawn accordingly to a discrete uniform distribution in $\{-1,1\}^d$, where $d$ denotes the data's dimension. Results on ACP and K are presented in Figures~\ref{fig:dim_ACP} and~\ref{fig:dim_K}, respectively.

We observe that \shortname outperforms \mshift and does slightly
better than \bms when looking at the ACP criterion, whereas for the K
parameter, its performance lies in between \mshift and \bms. All
algorithms provide satisfactory results in low dimensions, whereas for
higher dimensions, the \shortname could be preferred for better
results. As the data dimension increases and the number of
labels remains constant, the clustering operation is made easier
(leading to a high value of the ACP) whereas the ALP decreases
drastically along with K.

\subsubsection{Influence of the Number of Clusters}

\begin{figure*}[ht]
  \centering
  \begin{subfigure}{0.45\linewidth}
    \centering
    \includegraphics[width=0.9\linewidth]{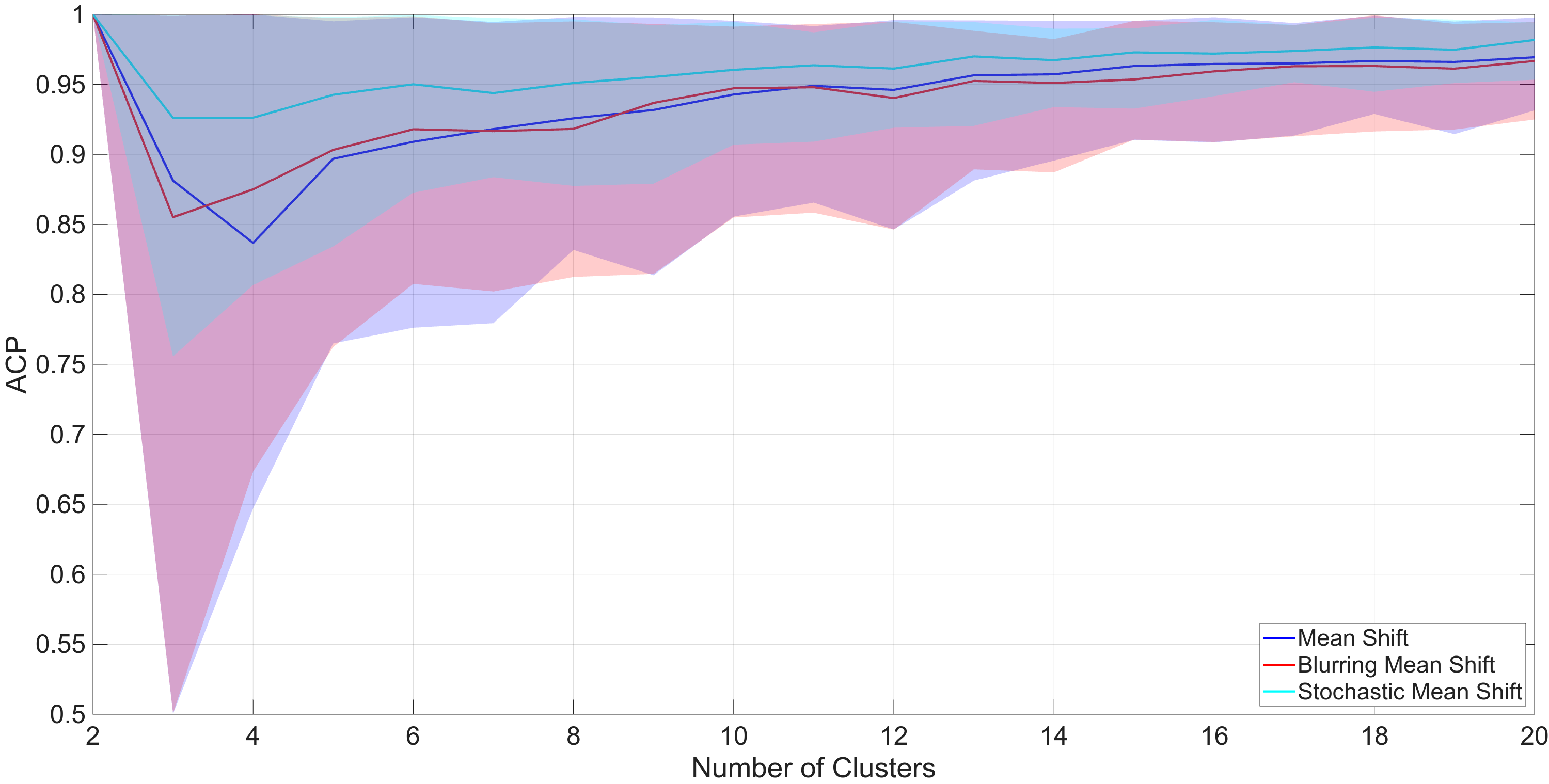}
    \caption{Influence on ACP.}
    \label{fig:nbc_ACP}
  \end{subfigure}
  \begin{subfigure}{0.45\linewidth}
  \includegraphics[width=0.9\linewidth]{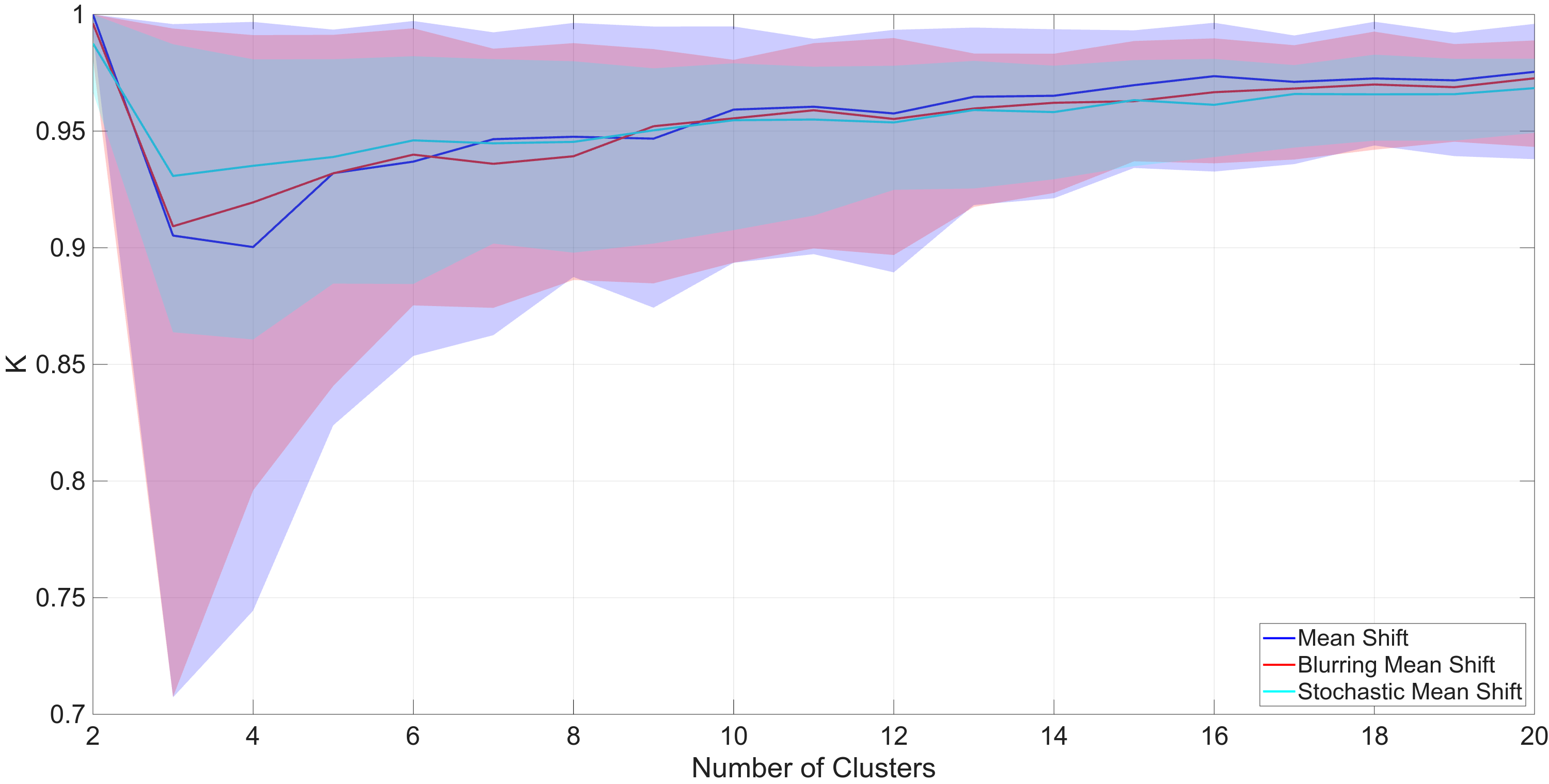}
  \caption{Influence on K.}
  \label{fig:nbc_K}  
  \end{subfigure}
  \caption{Influence of the number of clusters on clustering results for \mshift (blue), \bms (red) and \shortname (cyan).}
\label{fig:nbc}
\end{figure*}

Eventually, we investigate how the number of classes influences the clustering process. For this experiment, the Gaussian averages were drawn accordingly to a discrete uniform distribution in $\{-R/2, R/2\}^R$, where $R$ denotes the number of true classes. In some cases, this choice allows partial overlap between clusters, though overall the clustering task remains easy. Results on ACP and K are presented in Figures~\ref{fig:nbc_ACP} and~\ref{fig:nbc_K}, respectively.

Again, we observe that \shortname slightly outperforms other approaches, while the improvement becomes marginal as the number of clusters increases. We conclude that in most cases, the proposed algorithm is on par with the state of the art \bms and often behaves better than both deterministic approaches, particularly for unbalanced classes.


\section{Application to Speaker Clustering}
\label{sec:speech}

We present in this section an application of \shortname to speaker
clustering. 
The presented results were
developed in~\cite{Lapidot_2023}, and are reproduced in this paper for
convenience. 
%
Among other tasks, speaker clustering appears in speaker diarization~\cite{Aronowitz2020,Silnova2020,Lin2020}, which combines speaker segmentation as a first step and then applies a speaker clustering method. In this framework, the $ACP$ and the $ALP$ measures (in the original papers named \textit{average speaker purity} - $ASP$) as well as the $K$ value are commonly used to assess clustering validity~\cite{Ajmera2002,Salmun2017}.
Our results involve a large number of speakers, from $3$ to $60$ in
our simulations. The segments duration is about  $2.5 \, s $ on
average and the speaker segments are embedded as i-vectors.
We relied on the same database as
in~\cite{Salmun2016a,Salmun2016b,Salmun2017}, based on the NIST 2008
Speaker Recognition Database ~\cite{NIST2008_merged}, and used the
test corpus short2-short3-Test7 for male speakers only (198 speakers
in total). 
%
Mel Frequency Cepstral Coefficients (MFCC) and log energies were extracted from $25\,\rm ms$ frames (each one multiplied by a Hamming window), at the ratio of $19$ MFCCs 
and one log energy extracted per $10\,\rm ms$ segment. For each frame, 
 cepstral mean subtraction and variance normalization were performed on MFCCs, then delta and delta delta coefficients were calculated to finally yield a $60$-dimensional features vector. 


In the NIST 2008 database, the speech files consist of $5$ minutes phone speech. In the database we used, we already had i-vectors from an already segmented short speech segments. The distribution of the segment length $L$ has minimal length of $L_{min}=0.7\,\rm s$ and average length $L_{av}=2.5\,\rm s$. Its distribution can be approximated using an exponential distribution.
\begin{equation*}\label{SegmentLength}
\begin{array}{*{20}{c}}
{L \sim {L_{\min }} + \exp \left( \lambda  \right);} & {\lambda  = \frac{1}{{{L_{av}} - {L_{\min }}}}}
\end{array}
\end{equation*}
The average number of segments per speaker is of $\eta=34$ and standard deviation of $\sigma=6.0$. This distribution of the number of speakers $S$ can be approximated using Gaussian distribution, $S \sim {\cal N} \left( \eta,\sigma^2 \right )$.

Following~\cite{Salmun2016a,Salmun2016b}, the Probabilistic Linear
Discriminant Analysis (PLDA) scores matrix~\cite{Dehak2011} was used
to provide distance measures between points; at each step of both
\mshift and \shortname algorithms the neighborhood of any given  point
will be defined as the kNN neighborhood (with $k=17$ in our
application) w.r.t the PLDA scores related to this point.

We present the PLDA-based \shortname algorithm for speaker clustering
detailed in \cite{Lapidot2023}. In this framework, $\state = \left\{
  {{\point_j}} \right\}_{j = 1}^{n}$ is a set of normalized
i-vectors. The clustering algorithm includes three steps. First, a
preprocessing step is performed, which includes a principal component
analysis on the $\ell_2 $ normalized data $\left\{
  \widetilde{\point}_{j} := \point_j / \| \point_j \|\right\}_{j =
  1}^{n}$ to reduce data dimensionality from $d$ to $q<d$. We apply
further a whitening matrix $W$ to the projected data on the PCA subspaces  $\left\{ T(\widetilde{\point}_{j}) \right\}_{j = 1}^{n}$ and renormalize:
\begin{equation}
\label{SphericalNormalization}
\forall j \in [n]\,\ \, \varphi_{j}  = \frac{W \,
  T(\widetilde{\point}_{j})}{\| W \, T(\widetilde{\point}_{j}) \|} .
\end{equation}
Then, an integer $i \in [n]$ is drawn, and we find the subset
${S_k}\left(\varphi_i \right) \subseteq [n] $ containing the indices
$j$ having the $k$ highest PLDA scores $s\left( \varphi_j ,\varphi_i
\right)$, and we move the point $\point_i $ to its new location:
\begin{equation}
 \point_i \leftarrow \frac{\sum\limits_{j \in {S_k}\left( \varphi_i
     \right)} \point_j }{\#{S_k}\left( \varphi_i  \right)} .
\end{equation}
More details on the algorithm can be found in~\cite{Lapidot2023}.

The comparison between mean-shift based
speaker clustering algorithms, deterministic and stochastic, in terms
of $ACP$, $ALP$ and $K$, is shown in Table \ref{tab:diarizationResults}. In
addition, the \textit{average number of detected speakers}
(ANDS) is presented. Following \cite{Salmun2017} we compare the
algorithms for $3$, $7$, $15$, $22$, $30$ and $60$ speakers. Each
experiment was repeated $20$ times with randomly chosen speakers and randomly chosen segments per speaker.

\begin{table*}[!ht]
   \centering
\caption{Diarization results using \mshift and \shortname. For each value the result is given as \textbf{mean(std)}. The last column shows the number of times that \shortname achieved better performance than \mshift out of $20$ repetitions.}
   \maxsizebox{0.95\linewidth}{!}{\begin{tabular}{ccccccccccc}
  	\toprule
   \multirow{2}[3]{*}{\parbox[c]{0.2\linewidth}{\centering Speakers}} & \multicolumn{4}{c}{Deterministic Mean-Shift} & &\multicolumn{4}{c}{Stochastic Mean-Shift} & \\
   \cmidrule(l){2-5} \cmidrule(r){7-10} \\
    & \textbf{ACP} & \textbf{ALP} & \textbf{K} & \textbf{ANDS}  & \phantom{abcde} & \textbf{ACP} & \textbf{ALP} & \textbf{K} & \textbf{ANDS} & $SMS>MS$ \\
	\midrule
	$\mathbf{3}$ & $0.99\left(0.02\right)$ & $0.76\left(0.14\right)$ & $0.86\left(0.08\right)$ & $6.9\left(3.1\right)$ & & $0.97\left(0.09\right)$ & $0.87\left(0.12\right)$ & $0.91\left(0.08\right)$ & $7.0\left(2.5\right)$ & 13 \\

    $\mathbf{7}$ & $0.93\left(0.07\right)$ & $0.77\left(0.08\right)$ & $0.85\left(0.06\right)$ & $13.4\left(2.8\right)$ & & $0.92\left(0.10\right)$ & $0.85\left(0.08\right)$ & $0.88\left(0.07\right)$ & $16.1\left(3.1\right)$ & 16\\

	$\mathbf{15}$ & $0.92\left(0.04\right)$ & $0.75\left(0.06\right)$ & $0.83\left(0.04\right)$ & $26.4\left(2.7\right)$ & & $0.89\left(0.05\right)$ & $0.82\left(0.05\right)$ & $0.86\left(0.04\right)$ & $29.7\left(3.2\right)$ & 19\\

    $\mathbf{22}$ & $0.86\left(0.04\right)$ & $0.73\left(0.05\right)$ & $0.79\left(0.04\right)$ & $36.8\left(2.6\right)$ & & $0.83\left(0.06\right)$ & $0.79\left(0.05\right)$ & $0.81\left(0.04\right)$ & $41.9\left(4.2\right)$ & 13\\

	$\mathbf{30}$ & $0.82\left(0.04\right)$ & $0.68\left(0.05\right)$ & $0.75\left(0.04\right)$ & $46.8\left(5.2\right)$ & & $0.78\left(0.05\right)$ & $0.76\left(0.04\right)$ & $0.77\left(0.04\right)$ & $51.0\left(3.7\right)$ & 14 \\

    $\mathbf{60}$ & $0.75\left(0.03\right)$ & $0.67\left(0.03\right)$ & $0.71\left(0.03\right)$ & $78.9\left(4.2\right)$ & & $0.68\left(0.04\right)$ & $0.75\left(0.02\right)$ & $0.71\left(0.03\right)$ & $86.4\left(5.6\right)$ & 13\\
	\bottomrule
  \end{tabular}}
  \label{tab:diarizationResults}
\end{table*}


Overall, \shortname performs better than \mshift for $K$, except for the case of $60$ speakers where $K$ was the same. More precisely, the ALP in the stochastic case is significantly higher than in the deterministic case, whereas the ACP is slightly higher in the deterministic case. This result is surprising: typically \shortname produces more clusters than \mshift, suggesting i-vectors from one speaker might be split across more clusters, contrary to our findings. A possible explanation is that by isolating outliers into specific clusters, \shortname steps become less sensitive to their presence, resulting in an improved clustering adequacy overall. Note that \shortname was better than \mshift in $73.3\%$ of the cases ($88$ out of $120$ trials), and outperform MS independently of the number of speakers (based on the last column in Table~\ref{tab:diarizationResults}).

\section{Conclusion}
\label{sec:DiscussionConclusions}

In this work, we presented a stochastic version of the mean-shift clustering algorithm, \shortname. Results illustrating its convergence were presented. We investigated the robustness of \shortname in terms of number of clusters, dimensionality, computation times and class imbalance. Results showed that the proposed approach outperformed the ordinary Mean-Shift in most of the cases, and the more recent Blurring Mean-Shift in many of them. Furthermore, on the specific application of speaker clustering, we showed \shortname's potential in providing clusters of better quality than the commonly used Mean-Shift, which can be of great interest for practical applications. Further work is needed regarding the hyperparameters' optimal tuning. Moreover, this approach also raises the question of whether other hyperparameters could be drawn randomly for further improvements. These questions will be investigated in future contributions.







\bibliographystyle{elsarticle-num-names}
\bibliography{mybib}









\end{document}


\begin{frontmatter}
\title{Stochastic mean-shift clustering - Supplementary Material}


\author[1,2]{Itshak Lapidot}
\ead{itshakl@afeka.ac.il}
\address[1]{Afeka Tel-Aviv Academic College of Engineering, School of Electrical Engineering, Israel}
\address[2]{Avignon University, LIA, France}

\author[3]{Yann Sepulcre}
\ead{yanns@mail.sapir.ac.il}
\address[3]{Sapir Academic College, Department of Engineering, Sderot, Israel}

\author[4]{Tom Trigano}
\ead{thomast@sce.ac.il}
\address[4]{Shamoon College of Engineering, Department of Electrical Engineering, Ashdod, Israel}

\begin{abstract}
The document presents the proof of the paper ``Stochastic Mean-Shift Clustering''.
\end{abstract}

\end{frontmatter}  


\documentclass[./main.tex]{subfiles}

\begin{document}



\section*{Proof of Proposition~\ref{prop:non_decreasing_L}}

For any given $k \in \mathbb{N}$, we shall denote for simplicity $\state = \state^{(k)} = 
  [\point_{1}^T , \ldots , \point_{n}^{T} ]^{T} $ and $\nexts{\state} = \state^{(k+1)} = [\nexts{\point_{1}}^T , \ldots , \nexts{\point_{n}}^{T} ]^{T} $ ;
  introducing further $i=\ridx{k}$ as the index picked up randomly at step $k$, we have 
  $\forall \, j \in [n]\setminus \{i\} \, \, \nexts{\point_{j}} = \point_{j}$ so that 
  \begin{align}
      L(\nexts{\state}) - L(\state) & = k(0) + \sum\limits_{j \neq i } k \lpar \frac{\| \nexts{\point_{i}} - \point_{j}\|^2 }{h^2}\rpar - 
      \sum\limits_{j} k \lpar \frac{\| \point_{i} - \point_{j}\|^2 }{h^2}\rpar \label{eq:prop_step1}
  \end{align}
  In \eqref{eq:prop_step1} the sum in the last term is on every $j \in \neighb{i}$; since $k$ has only non-negative values one can pursue as follows:
  \begin{align}
      L(\nexts{\state}) - L(\state) & \geq \sum\limits_{\substack{j \in \neighb{i} \\ j \neq i}} k \lpar \frac{\| \nexts{\point_{i}} - \point_{j}\|^2 }{h^2}\rpar - k \lpar \frac{\| \point_{i} - \point_{j}\|^2 }{h^2}\rpar \nonumber \\
      & \geq \frac{1}{h^2 } \sum\limits_{\substack{j \in \neighb{i} \\ j \neq i}} G\lpar \frac{\point_{i} - \point_{j}}{h}\rpar \lpar \| \point_{i} - \point_{j}\|^2 -  \| \nexts{\point_{i}} - \point_{j}\|^2 \rpar \label{eq:prop_step2} \\
      & =  \frac{1}{h^2 } \lbr G(\mathbf{0}) \| \nexts{\point_{i}} - \point_{i}\|^2 + 
     \sum\limits_{j \in \neighb{i}}  G\lpar \frac{\point_{i} - \point_{j}}{h}\rpar \lpar \| \point_{i} - \point_{j}\|^2 -  \| \nexts{\point_{i}} - \point_{j}\|^2 \rpar\rbr \nonumber 
  \end{align}
  In \eqref{eq:prop_step2} the inequality follows from the convexity assumption on $k$: indeed for every $(\mathbf{u}, \mathbf{v}) \in \lpar \re^n \rpar^2 $ we have $k(\| \mathbf{v}\|^2 ) - k (\| \mathbf{u}\|^2 ) \geq k^{\prime} (\| \mathbf{u}\|^2 ) \lpar \| \mathbf{v}\|^2 - \| \mathbf{u}\|^2 \rpar = G(\mathbf{u}) \lpar \| \mathbf{u}\|^2 - \| \mathbf{v}\|^2 \rpar $. In the last expression above, by the very definition
  of $\nexts{\point_i }$ as a mean point: $\dsp \nexts{\point_i } = \sum\limits_{j \in \neighb{i}} \frac{\dsp G \lpar \frac{\dsp \point_i - \point_j }{h}\rpar }{\dsp \sum\limits_{m \in \neighb{i}} G \lpar \frac{\point_i - \point_m }{h}\rpar }  \point_j $ , we have the simple equality 
  \begin{align}
  \sum\limits_{j \in \neighb{i}}  G\lpar \frac{\point_{i} - \point_{j}}{h}\rpar \lpar \| \point_{i} - \point_{j}\|^2 -  \| \nexts{\point_{i}} - \point_{j}\|^2 \rpar & = \lpar \sum\limits_{j \in \neighb{i}} G \lpar \frac{\point_i - \point_j }{h}\rpar \rpar  
  \| \point_i - \nexts{\point_{i}} \|^2 \, \, ; \nonumber 
  \end{align}
  it follows immediately by the non-negativity of $G$ that 
  \begin{align}
      L(\nexts{\state}) - L(\state) & \geq \frac{2 \, G(\mathbf{0})}{h^2 }  \| \nexts{\point_{i}} - \point_{i}\|^2 \label{eq:prop_stepfinal}
  \end{align}
  In turn \eqref{eq:prop_stepfinal} implies that $\lpar L(\state^{(k)})\rpar_{k \in \mathbb{N}} $ is a non decreasing, hence convergent sequence since it is upper bounded by the constant $\frac{n(n+1)}{2}\, k(0)$. 
  As for the second claim, it follows easily from the equality 
  $ \nabla_{i } L (\state^{(k)}) = \frac{2 \sum\limits_{j=1}^n G\lpar \frac{\point_{i}^{(k)} - \point_{j}^{(k)}}{h}\rpar}{h^2 } \lbr \state^{(k+1)} - \state^{(k)} \rbr $ hence 
  \begin{align}\label{eq:bounding_partial_gradient_SBMS}
  \| \nabla_{i } L (\state^{(k)}) \| \leq  \frac{2\,n  \, |k^{\prime}(0)|}{h^2 } \| \nexts{\point_{i}} - \point_{i} \| 
  \end{align}
   since by assumption $G(\mathbf{u})=|k^{\prime}(\| \mathbf{u}\|^2 )|$ is a non increasing function of the norm $\| \mathbf{u} \|$. 
  Now 
  \eqref{eq:prop_stepfinal}\eqref{eq:bounding_partial_gradient_SBMS} lead immediately to the following
  \begin{align}\label{eq:prop_end_proof}
      \| \nabla_{i } L (\state^{(k)}) \| & \leq \frac{n \, \sqrt{2 \, | k^{\prime}(0)|}}{h}\, \lpar L(\nexts{\state}) - L(\state) \rpar^{1/2}   
  \end{align}
  which completes the proof.

\section*{Proof of Lemma \ref{lem:preliminary_to_prop_on_stopping_times}}

 It is clear that $\Omega_{p+1} = \bigcup\limits_{k=k_0 + p+1}^{\infty} \lpar T_{p+1} = k \rpar $. The probability $\prob{\Omega_{p+1}}$ can be decomposed as follows
\begin{align}
    \prob{\Omega_{p+1}} & =  \sum\limits_{k=k_0 +p+1}^{\infty} 
    \sum\limits_{m=k_0 +p}^{k-1} \prob{T_{p+1} = k \ \wedge \ T_{p}=m } \nonumber \\
    & = \sum\limits_{k=k_0 +p+1}^{\infty} 
    \sum\limits_{m=k_0 +p}^{k-1} \mathbb{P} ( \, \| \nabla^{(k)} \| \geq \epsilon \ 
    \wedge \ \| \nabla_{\ridx{k}}^{(k)} \| < \epsilon \ \cdots \nonumber \\
   & \hspace{3cm}  \ \wedge \ \forall \, j \in (m , \ldots , k) \  \| \nabla^{(j)} \| < \epsilon 
    \ \wedge \ T_{p}=m ) \label{eq:proof_lemma_step1}
    \end{align}
    Consider any given pair of integers $(k,m)$ such that 
    $k\geq k_0 +p+1$ and $k_0 + p \leq m \leq k-1$, and define the following event as a subset in $[n]^{k-1}$:
    \begin{align}
    B_{k,m} = \lset \lbr \ridx{k-1},\ldots,\ridx{0}\rbr \, | \ \| \nabla^{(k)} \| \geq \epsilon \ \wedge \ \forall \, j \in (m , \ldots , k) \  \| \nabla^{(j)} \| < \epsilon  \ \wedge \ T_{p}=m \rset  \label{def:event_B}
    \end{align}
    As was mentioned previously right after \eqref{eq:step_sbms}: at any step $k$ of \shortname the state $\state^{(k)}$ depends solely on the random sequence of indices $[\ridx{k-1},\ldots,\ridx{0}]$ ; hence there exists a (non random) function $\mathcal{E}^{(k)}: [n]^{k} \to 2^{[n]}$ such that 
    $\lset i \in [n] \, | \ \| \nabla_{i}^{(k)} \| < \epsilon \rset = \mathcal{E}^{(k)} \lbr \ridx{k-1},\ldots,\ridx{0} \rbr $. 
     As a consequence, using the assumption that the index
    $\ridx{k}$ is independent of $\lbr \ridx{k-1},\ldots,\ridx{0}\rbr $
    it holds that 
    \begin{align}
        \prob{\| \nabla_{\ridx{k}}^{(k)} \| < \epsilon \ \wedge \ B_{k,m}} & = 
        \sum\limits_{\lbr \ridx{k-1},\ldots,\ridx{0}\rbr \in B_{k,m}} 
        \prob{\ridx{k} \in \mathcal{E}^{(k)} \lbr \ridx{k-1},\ldots,\ridx{0} \rbr }
        \prob{\lbr \ridx{k-1},\ldots,\ridx{0}\rbr} \nonumber \\
        & \leq \frac{n-1}{n} \prob{B_{k,m}} \label{eq:proof_lemma_step2} \ \ ; 
    \end{align}
    last inequality \eqref{eq:proof_lemma_step2} is satisfied since for all 
    $\lbr \ridx{k-1},\ldots,\ridx{0}\rbr \in B_{k,m} $ it holds that 
    $\| \nabla^{(k)} \| \geq \epsilon $ hence $\exists \, i \ \| \nabla_{i}^{(k)} \| \geq \epsilon$, implying the rough upper bound on the cardinality
    $\# \mathcal{E}^{(k)} \lbr \ridx{k-1},\ldots,\ridx{1} \rbr \leq n-1$. Now going back to \eqref{eq:proof_lemma_step1} one can write:
    \begin{align}
     \prob{\Omega_{p+1}} & \leq \frac{n-1}{n} \sum\limits_{m=k_0 +p}^{\infty}
    \sum\limits_{k=m+1}^{\infty} \prob{B_{k,m}} \nonumber \\
    & \leq \frac{n-1}{n} \sum\limits_{m=k_0 +p}^{\infty} \prob{T_{p}=m} = \lpar 1 - \frac{1}{n}\rpar \prob{\Omega_{p}} \label{eq:proof_lemma_step3} \ \ ; 
\end{align}
 indeed last inequality \eqref{eq:proof_lemma_step3} follows from the immediate 
 $\sum\limits_{k=m+1}^{\infty} \prob{B_{k,m}} \leq \prob{T_{p}=m} $ which holds for any $m \in \nn $.

\section*{Proof of Proposition \ref{prop:nabla_goes_to_zero_as}}

Given any $\epsilon >0$ , by proposition \ref{prop:non_decreasing_L} for any $\omega = [\ridx{k}]_{k \in \nn}$ and $\epsilon>0$ it holds a.s that 
$ \exists \, K (\omega) \in \nn \ \, \forall \, k > K (\omega) \ \, 
    \| \nabla_{\ridx{k}}^{(k)} \| < \epsilon $. For any $k_0 \in \nn $ let us define the event $\mathcal{U}_{k_0 } = \{ K (\omega) = k_0 \}$ ; lemma 
    \ref{lem:preliminary_to_prop_on_stopping_times} shows that for the sequence of stopping times $(T_p )$ defined with the pair $(\epsilon , k_0 )$ by \eqref{def:stopping_1} \eqref{def:stopping_p} it holds that
    $\lim\limits_{p \to \infty} \prob{T_p < \infty \ \wedge \ \mathcal{U}_{k_{0}}} = 0$, in other words $\prob{\lpar \forall \, p \in \nn \ T_p < \infty\rpar  \, \wedge \, \mathcal{U}_{k_{0}}} = 0$.
    Since $\prob{\bigcup\limits_{k_0 \in \nn} \mathcal{U}_{k_0 }}=1$ by proposition \ref{prop:non_decreasing_L}, it follows immediately that 
    $\prob{\lpar \forall \, p \in \nn \ T_p < \infty\rpar} = 0$, in other words there is a.s. on $\omega $ an integer $K_1 (\omega)$ such that 
    $\forall \, k \geq K_1 (\omega) \ \, \| \nabla^{(k)} (\omega) \| < \epsilon$. The result is proved.

\section*{Proof of Theorem \ref{thm:clusters_existence}}

  Consider the following event $\mathcal{B}$ (in the condition below $(k_p )$ denotes an infinite integers sequence): 
  \begin{align}
      \mathcal{B} = \lset \omega \,| \ \exists \, (k_p) \, \ \forall \, p \in \nn \, \, \exists (i,j) \in [n]^2 \ 
      \tau \leq \| \point_{i}^{(k_p )} - \point_{j}^{(k_p )} \| \leq h-\tau \rset \nonumber
  \end{align}
  
  For any $\omega \in \mathcal{B}$, considering a subsequence of states $
  \lbr \state^{(k_p )} \rbr $ satisfying the condition above, 
  clearly there exist two distinct and fixed integers $i\neq j $ in $[n]$ such that for an infinity of integers $(k_{p_m})$ one has
 $ \tau \leq \| \point_{i}^{(k_{p_m} )} - \point_{j}^{(k_{p_m} )} \| \leq h-\tau $ for any $m \in \nn$. By compactness one can further extract some  subsequence of $\lbr \state^{k_{p_m}} \rbr $ which tends to some $ \overline{\state} \in \re^{d n}$
  (indeed, for any $k \in \nn $ all the points $\point_{i}^{(k)}$ belong to the convex envelop $\text{conv}\lpar \point_{i}^{(0)}, \, i \in [n]\rpar$ which is compact); but it would imply that $ \tau \leq \| \overline{\point}_{i} - \overline{\point}_{j} \| \leq h-\tau $, hence $\nabla L (\overline{\state}) \neq \mathbf{0}$ by theorem \ref{thm:critical_point_charact}. 
  By our assumption $L$ is $C^1$, so the previous argument shows the inclusion $\mathcal{B} \subseteq \{ \omega \, | \, \nabla^{(k)} \not\to \mathbf{0}_{d n}\}$, and since the latter subset is of measure $0$ by proposition \ref{prop:nabla_goes_to_zero_as} it follows that $\prob{\mathcal{B}}=0$. 

  Therefore, it holds almost surely on $\omega $ that there exists
  $K= K(\omega) \in \nn $ such that the following two conditions hold 
  \begin{align}
  \forall \, k \geq K &  \ \ \, \forall \, (i,j) \in [n]^2 \ \, \| \point^{(k)}_{i} - \point_{j}^{(k)} \| < \tau \ \, \vee \ \, \| \point^{(k)}_{i} - \point_{j}^{(k)} \| > h- \tau \label{thm_pf_cond1} \\
   \forall \, k \geq K & \ \ \, \| \nabla^{(k)} \| \leq \frac{2 G(\mathbf{0})}{h^2 } \lpar h - 2 \tau \rpar  \label{thm_pf_cond2} ;
  \end{align}
  suppose now that two integers $(i,j)\in [n]^2 $ satisfies for some $k \geq K$ that $\| \point^{(k)}_{i} - \point_{j}^{(k)} \| < \tau $ and
  $\| \point^{(k+1)}_{i} - \point_{j}^{(k+1)} \| > h- \tau $ (or these two inequalities in opposite order, the argument will stay the same); since at any step \shortname modifies one point only, we may assume that $\point_{j}^{(k+1)}=\point_{j}^{(k)}$ and therefore
  \begin{align}
      \| \point^{(k+1)}_{i} - \point^{(k)}_{i} \| & \geq \| \point^{(k+1)}_{i} - \point_{j}^{(k+1)} \| - \| \point^{(k)}_{i} - \point_{j}^{(k)} \| \nonumber \\
      & >  h-2 \tau 
  \end{align}
  which contradicts this other property which follows directly from 
  \eqref{eq:step_sbms} and assumption \eqref{thm_pf_cond2} above:
  $\| \point^{(k+1)}_{i} - \point^{(k)}_{i} \| \leq \frac{h^2 }{2 G(\mathbf{0})} \| \nabla^{(k)} \| \leq h - 2 \tau$. In other words 
  conditions \eqref{thm_pf_cond1}\eqref{thm_pf_cond2} above imply the existence 
  of some (fixed) partition $J_1 ,\ldots,J_M $ of $[n]$ satisfying for every $k \geq K$ conditions $1)2)$ in the theorem. Since $\tau$ was arbitrary positive the diameter of each cluster $\mathcal{C}_{l}, \, l \in [M]$ tends a.s to $0$ as $k \to \infty$.

\section*{Proof of Theorem \ref{thm:convergence_for_one_cluster}}

  For any finite set $S \subset \re^d $ of points, we shall denote by $\text{conv}(S)$ the convex hull of $S$ which in this case is a compact set; since for any $k \in \nn$ we have 
  $\text{conv}\lpar \state^{(k+1)} \rpar \subseteq \text{conv}\lpar \state^{(k)}\rpar $, it follows that $\exists \, \mathbf{z} \in \bigcap\limits_{k \in \nn} \text{conv}\lpar \state^{(k)}\rpar $; by defining the transformed state
  $\widehat{\state}^{(0)} = \lbr (\point_{1}^{(0)} - \mathbf{z})^T , \ldots , 
  (\point_{n}^{(0)} - \mathbf{z})^T \rbr^T $ we shall consider from now that $\mathbf{z}=\mathbf{0}_d $, with the algorithm proceeding as usual with $\widehat{\state}^{(0)}$ as the original state. In this case we shall prove that for any $i \in [n]$: $\lim\limits_{k \to \infty} \point_{i}^{(k)} = \mathbf{0}_{d}$.

  Let $\mathbf{u} \in \re^d $ be any unit vector, and denote by $\pi_{\mathbf{u}} : \re^d \to L_{\mathbf{u}} $ the orthogonal projection operator on the line 
  $L_{\mathbf{u}} = \re \mathbf{u}$. For any $k \in \nn $ we  define the projected points $y_{i}^{(k)} = \pi_{\mathbf{u}} (\point_{i}^{(k)})$ on $L_{\mathbf{u}}$ for any $i \in [n]$, and consider these as $n$ real numbers. By the linearity of $\pi_{\mathbf{u}}$ the sequence of states 
  $\lpar \lbr y_{1}^{(k)} , \ldots , y_{n}^{(k)} \rbr \rpar_{k \in \nn} $ (here a state is a point in $\re^n $) is updated by the rule:
  \begin{align} \label{eq:weighted_average} i=\ridx{k}\, : \ \ y_{i}^{(k+1)} =  \frac{\sum\limits_{j \in [n]} \dsp G \lpar \frac{\dsp \point_{i}^{(k)} - \point_{j}^{(k)} }{h}\rpar y_{j}^{(k)} }{\dsp \sum\limits_{j \in [n]} G \lpar \frac{\point_{i}^{(k)} - \point_{j}^{(k)} }{h}\rpar } \ \ ; \end{align}
  note that the summations above are on all $j \in [n]$ since $\neighb{i}=[n]$. 
  For any $k \in \nn$ define $m^{(k)} = \min \lset y_{i}^{(k)} , \, i \in [n] \rset\, , \, M^{(k)} = \max \lset  y_{i}^{(k)} , \, i \in [n] \rset $; the sequence $\lpar m^{(k)}\rpar $ (resp. $ \lpar M^{(k)}\rpar $) is non decreasing (resp. non increasing), hence both sequences converge to real numbers:
  $\lim\limits_{k \to \infty} m^{(k)} = m \leq M = \lim\limits_{k \to \infty}M^{(k)}$. We now show that $m=M$ holds a.s, or in other words 
  that $\prob{\lset m<M\rset }=0$. Suppose $\omega $ satisfies $ m(\omega) < M(\omega)$, and let $\epsilon >0 $ be any positive value which shall be chosen conveniently below; so there exists an integer $K_0$ such that for all $k \geq K_0 $ we have $m-\epsilon < m^{(k)}$. Given any such $k \geq K_0 $, denote by $i_k$ the chosen integer, and let $j_k \in [n]$ be some integer such that $y_{j}^{(k)} \geq M$ (there has to be one at least); we denote also the positive weights in \eqref{eq:weighted_average} simply as follows:
  $ \forall \, p \in [n] \ \, w_p := G \lpar \frac{\dsp \point_{i}^{(k)} - \point_{p}^{(k)} }{h}\rpar $. It holds that 
  $$ \forall \, k \geq K_0 \ \ y_{i_k }^{(k+1)} > \frac{\sum\limits_{p \neq j_k } w_p (m-\epsilon) + w_{j_k} \, M}{\sum\limits_{p \neq j_k } w_p + w_{j_k} } \ \ \, ; $$
  now for $\epsilon$ chosen sufficiently small the quantity on the RHS of the previous inequality is greater than $m$, as this property is equivalent to 
  $w_{j_k} (M-m) > (\sum\limits_{p \neq j_k } w_p) \epsilon $ which is guaranteed
if we choose $\dsp\epsilon < \frac{  \left | k^{\prime} \lpar \frac{\text{diam}(\state^{(0)})^2}{h^2} \rpar \right |}{(n-1)|k^{\prime}(0)|} (M-m) $, since $\text{diam}(\state^{(k)})$ is a decreasing sequence. So far we have proved that for any $\omega $ such that $m(\omega)<M(\omega)$ there is a $K_0 $ (depending on $\omega$) such that $\forall \, k \geq K_0 \ \, y_{i_k }^{(k+1)} > m$; since the indices $i_k \sim \mathcal{U}_{[n]}$ are i.i.d every $i \in [n]$ will be picked at least once within a finite number of steps after $K_0$ a.s., such event implying that $\exists \, K_1 \  m^{(K_1)}>m$, a contradiction by the monotonicity of $m^{(k)}$.
 Since $K_0 \in \mathbb{N}$, the event $\{ m<M \}$ is expressed as a denumerable union of null measure subsets, hence $\prob{\lset m<M\rset }=0$ as claimed. 
 
  By assumption $\forall \, k \in \nn \ \, 0 \in \text{conv}\lpar y_{i}^{(k)} \, , \ i\in [n] \rpar = \lbr m^{(k)} , M^{(k)} \rbr $, so the fact that $\lim\limits_{k \to \infty} M^{(k)} - m^{(k)} =0 $ holds a.s. implies that for all $\mathbf{u} \in \re^d $ and 
  for each $i \in [n]$ one has $\lim\limits_{k \to \infty} \pi_{\mathbf{u}} (\point_{i}^{(k)}) = 0 $ a.s.
  The same argument can be used for any unit vector $\mathbf{u} \in \re^d $, so the theorem is proved.

\end{document}


\makeatletter\@input{xx.tex}\makeatother